\documentclass[lettersize,journal]{IEEEtran}
\usepackage{amsmath,amsfonts}
\usepackage{algorithmic}
\usepackage{array}
\usepackage[caption=false,font=normalsize,labelfont=sf,textfont=sf]{subfig}
\usepackage{textcomp}
\usepackage{stfloats}
\usepackage{url}
\usepackage{verbatim}
\usepackage{graphicx}
\usepackage{booktabs}
\usepackage{multirow}
\usepackage{booktabs}
\usepackage{caption}
\usepackage{float} 
\usepackage{tabularx}
\usepackage{enumitem}
\usepackage{hyperref}
\usepackage{orcidlink} 
\hypersetup{hidelinks} % hide the hyperlink box
\usepackage{xcolor} % 引入颜色宏包

\def\BibTeX{{\rm B\kern-.05em{\sc i\kern-.025em b}\kern-.08em
    T\kern-.1667em\lower.7ex\hbox{E}\kern-.125emX}}
\usepackage{balance}
\begin{document}
%yuyinfeng orcid :\orcid{0000-0003-3089-4140}
%wang liejun:0000-0003-0210-2273
\title{FSDENet: A Frequency and Spatial Domains based Detail Enhancement Network for Remote Sensing Semantic Segmentation}
\author{Jiahao Fu, Yinfeng Yu$^{\dagger}\orcidlink{0000-0003-3089-4140}$, ~\IEEEmembership{Member,~IEEE,} and Liejun Wang$^{\dagger}\orcidlink{0000-0003-0210-2273}$ 
\thanks{$^{\dagger}$Both Yinfeng Yu and Liejun Wang are corresponding authors.}
\thanks{This study was funded by the Excellence Program Project of Tianshan, Xinjiang Uygur Autonomous Region, China (grant number 2022TSYCLJ0036), the Central Government Guides Local Science and Technology Development Fund Projects (grant number ZYYD2022C19), and the National Natural Science Foundation of China (grant numbers 62472368, 62463029, and 62303259).
The authors are with the School of Computer Science and Technology,
Xinjiang University, Urumqi 830046, China (e-mail: 107552301345@stu.xju.edu.cn;yuyinfeng@xju.edu.cn;wljxju@xju.edu.cn).
}
}

\markboth{}%
{}

\maketitle

\begin{abstract}
To fully leverage spatial information for remote sensing image segmentation and address semantic edge ambiguities caused by grayscale variations (e.g., shadows and low-contrast regions), we propose the Frequency and Spatial Domains based Detail Enhancement Network (FSDENet). Our framework employs spatial processing methods to extract rich multi-scale spatial features and fine-grained semantic details. By effectively integrating global and frequency-domain information through the Fast Fourier Transform (FFT) in global mappings, the model's capability to discern global representations under grayscale variations is significantly strengthened. Additionally, we utilize Haar wavelet transform to decompose features into high- and low-frequency components, leveraging their distinct sensitivity to edge information to refine boundary segmentation. The model achieves dual-domain synergy by integrating spatial granularity with frequency-domain edge sensitivity, substantially improving segmentation accuracy in boundary regions and grayscale transition zones. Comprehensive experimental results demonstrate that FSDENet achieves state-of-the-art (SOTA) performance on four widely adopted datasets: LoveDA, Vaihingen, Potsdam, and iSAID.
\end{abstract}


\section{Introduction}
\IEEEPARstart{T}{he} continuous advancement of sensor technology, in conjunction with the rapid development of the aerospace field, has resulted in the increasing accessibility of high-resolution satellite and aerospace remote sensing images. These images provide detailed documentation of various geographical landscapes, including urban buildings, farmland, forests, and lakes. Consequently, high-resolution remote sensing data is increasingly available for scientific research and practical applications. Remote sensing image segmentation techniques, as a key method for subdividing images of the Earth's surface into different objects or classes, play a crucial role in numerous domains,  including geographic information systems (GIS), agricultural planning\cite{Agriculture}, land change\cite{Land1}\cite{land2}, environmental monitoring\cite{climate}, and crisis management\cite{Crisismanagement}.

In recent years, deep learning—particularly Convolutional Neural Network (CNN)—has achieved remarkable breakthroughs in the semantic segmentation of natural images. Representative methods such as FCN\cite{Fcn}, UNet\cite{Unet}, and DeepLabV3+\cite{DeeplabV3++} have been widely adopted in domains like medical imaging and autonomous driving due to their powerful feature extraction and representation capabilities. However, directly applying these approaches to remote sensing imagery presents several challenges. Remote sensing images typically exhibit higher resolutions, more complex background textures, significant scale variations, a higher density of small objects, and interference factors such as shadow occlusions. These characteristics limit the ability of CNN, which possesses inherently restricted receptive fields, to capture global semantic context, thereby reducing accuracy in recognizing fine edges and small objects.

To address these challenges, extensive research has focused on enhancing CNN architectures to improve their adaptability to remote sensing data. For example, HRNet\cite{HRNet} maintains high-resolution feature maps through multi-branch structures, thereby preserving fine-grained details. The DeepLab series \cite {DeeplabV3++} utilizes atrous convolution to expand the receptive field. Architectures like UNet\cite{Unet} fuse multi-scale feature information through dense and skip connections, effectively retaining low-level edge features. FarSeg++\cite{Farseg++} incorporates a foreground enhancement mechanism to improve the perception of small objects. Meanwhile, the Transformer architecture, which leverages self-attention mechanisms, has demonstrated strong global modeling capabilities in works such as ViT\cite{ViT} and Swin Transformer\cite{Swin}, breaking the limitations of local receptive fields. In the remote sensing domain, methods such as Segmenter\cite{Segmenter} and SwinUNet\cite{swinunet} further advance pure Transformer-based architectures, thereby enhancing semantic understanding in complex scenes.

Despite these advancements in accuracy, Transformers still face key limitations—specifically, the computational complexity of multi-head self-attention scales quadratically with image size, making them unsuitable for ultra-high-resolution satellite imagery. To mitigate this issue, a growing body of research explores hybrid CNN-Transformer architectures\cite{jstart1}. For instance, ConvLSR-Net\cite{LSRFormer} integrates lightweight convolutional modules for local feature extraction and enhances global modeling through Transformer blocks. UnetFormer\cite{Unetformer} embeds Swin Transformer modules within the decoder, enabling effective fusion of local and global information. CMTFNet\cite{cmtfnet} proposes a multi-scale Transformer fusion strategy to model cross-scale semantic relationships at various levels. These hybrid approaches successfully address the respective limitations of CNNs and Transformers, achieving excellent performance in remote sensing image segmentation tasks.

\begin{figure}[t]
    \centering
    \includegraphics[width=\linewidth]{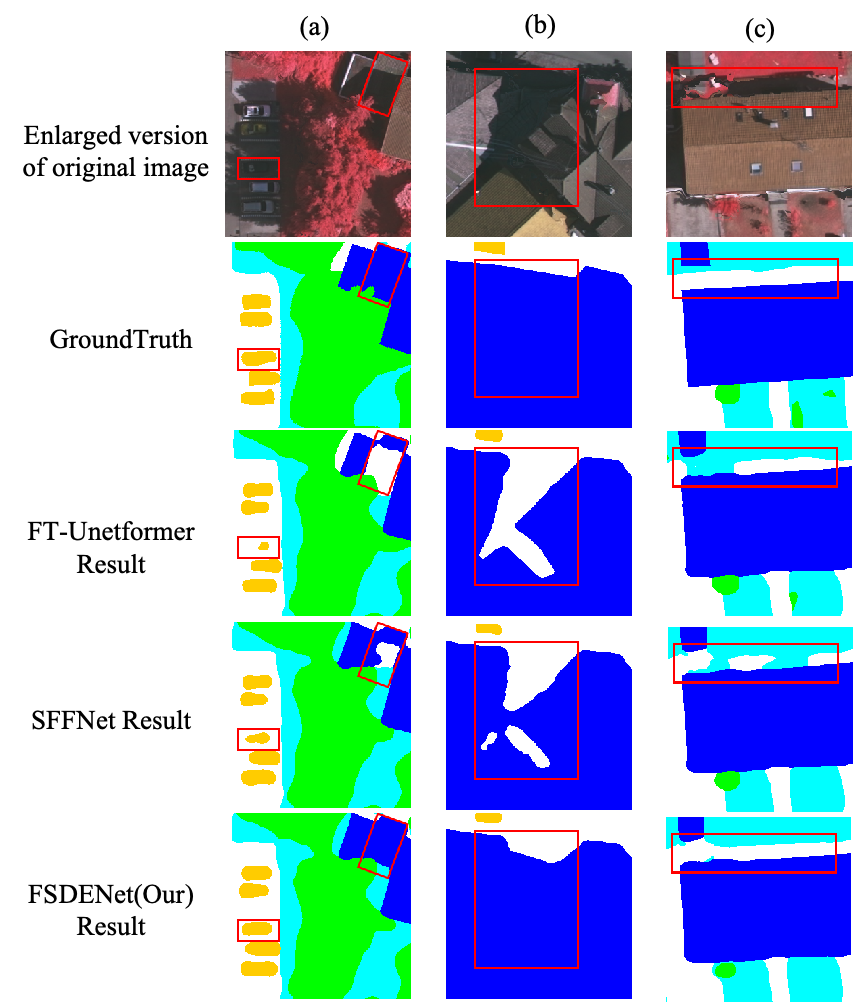}
    \captionsetup{font={small}}
    \caption{The figure illustrates the current challenges of remote sensing image segmentation: facing regions with large grey scale changes, such as shadows and low-contrast regions with obvious semantic ambiguities, it isn't easy to segment accurately. The first line is a local zoomed version of the original image, the second line corresponds to ground-truth labels (GT), the third line is the FT-UNetFormer segmentation result, the fourth line is the SFFNet segmentation result of the latest SOTA method, and the fifth line is the FSDENet segmentation result. It can be seen from the results that FT-UNetFormer, which only uses spatial information, performs poorly in dealing with shaded, low-contrast regions (e.g., the car is obscured by shadows, causing the low-contrast boundary to be inconspicuous). SFFNet, which adds frequency-domain information, significantly improves such problems. Our method makes full use of frequency-domain information to achieve better results.}
    \label{fig:architecture}
\end{figure}

% Multi-scale features are also vital in processing high-resolution remote sensing images, intending to increase the accuracy of the segmentation of objects of different scales and small target objects. However, it should be noted that a single pixel in a deep feature originates from a pixel region in a shallow feature. For example, an agriculture or lake area in a shallow feature is represented as a single pixel point in a deep feature. Simple addition or concatenation of multi-scale features is inadequate in addressing this issue of mismatched receptive fields. This can lead to problems such as shallow edge texture information being overwhelmed by deep semantic information, which can compromise the accuracy of edge segmentation, particularly in the context of semantic segmentation of high-resolution remote sensing images.	

Although both CNN and Transformer-based methods have achieved significant progress in semantic segmentation of remote sensing images, most current approaches still rely primarily on spatial-domain feature modeling, often overlooking the rich frequency-domain information inherent in such data\cite{Xnet}. In practice, remote sensing imagery frequently contains shadow occlusions, low-contrast regions, and texture-blurred boundaries—features that are typically reflected in the high-frequency components of the frequency domain\cite{WaveConv}. Traditional spatial-domain methods inherently struggle to represent these frequency-sensitive features effectively. Since frequency-domain information is susceptible to grayscale variations, its proper utilization can significantly enhance a model’s ability to perceive boundaries and fine details. In recent years, SFFNet\cite{SFFnet} made an initial attempt to integrate frequency features extracted via Haar wavelet transform into spatial feature representations, yielding promising results. However, this approach provides only a preliminary exploration of frequency-domain features, lacking deeper global frequency modeling and dedicated strategies for detail enhancement.
% It is important to note that noise and low-contrast regions frequently exist in remote sensing images, such as edges, shadows, or small-scale changes (contours of trees, etc.). These regions correspond to the high-frequency part of the frequency domain, and noise often appears in remote sensing images in the high-frequency part of the frequency domain. These regions are also the places where segmentation is usually prone to errors. The texture and edge features of the target in remote sensing images are expressed as specific frequency components in the frequency domain. The extreme sensitivity of the frequency domain to grey-scale changes\cite{SFFnet} means that frequency domain analysis can supplement the information win the frequency domain, effectively filter out the noise, and simultaneously enhance the saliency of the target area. This enables better modeling of global and local properties of remote sensing images and improves the model's boundary segmentation.

We adopt a UNet-like architecture in which the fusion of shallow and deep features facilitates enhanced information flow from the early to later stages of the network. While traditional feature fusion methods typically rely on simple addition or weighted summation operations \cite{Unetformer}, it is important to recognize that a single pixel in a deep feature map often corresponds to a broader region in the shallow feature map. For example, an area representing farmland or a lake in the shallow features may be compressed into a single pixel in the deeper layers. The direct addition or concatenation of multi-scale features does not adequately address this mismatch in receptive fields \cite {DEA}. As a result, fine-grained edge and texture information from shallow layers can be overwhelmed by the semantic abstraction present in deeper layers, ultimately degrading edge segmentation accuracy \cite{FFusion}. To mitigate this issue, we propose a feature fusion strategy that leverages both hybrid channel attention and spatial attention mechanisms, enabling the adaptive integration of low-level encoder features with their corresponding high-level counterparts.

% \textcolor{blue}{Inspired by the aforementioned principles, we propose FSDENet, a spatial-frequency fused detail enhancement network for remote sensing image segmentation. First, to better preserve edge and texture details, this work first utilizes Convolutional Neural Networks (CNNs) to extract local multi-scale features, providing rich detail information for subsequent feature fusion. The Mulit-Attention Select Fusion Block (MASF) addresses the receptive field mismatch by aligning multi-scale features using hybrid attention, ensuring shallow edge features are preserved. To compensate for CNN’s limited global context modeling, we introduce the Cross Agent-Attention Global Filter (CAGF) for global semantic aggregation. Furthermore, the ast Fourier Detail Perception Block (FFDP) leverages Fast Fourier Transform (FFT) to extract global frequency features, enhancing sensitivity to grayscale variations. Finally, the Haar Wavelet transform Detail Enhancement Block (HDEE) employs Haar wavelet transform to decompose spatial features into high- and low-frequency components, applying targeted enhancements to boost boundary perception. Together, these modules achieve effective spatial-frequency fusion, significantly improving segmentation performance in challenging conditions such as shadows, weak edges, and low-contrast regions.}

Based on these insights, we propose FSDENet, a frequency and spatial domains-based detail enhancement network designed to comprehensively improve edge awareness and robustness to grayscale variation in remote sensing semantic segmentation. Specifically, our main contributions are as follows:
\begin{enumerate}
    \item We design a Mulit-Attention Select Fusion Block (MASF) that integrates spatial and channel attention. By explicitly learning the importance of spatial locations, the module guides channel-wise feature modulation to preserve fine-grained edge and texture details in shallow layers. This design effectively mitigates the suppression of structural information by deep semantic features.
    
    \item We design a Cross Agent-Attention Global Filter (CAGF) to address the difficulty of convolutional structures in capturing global dependencies. This module enables efficient inter-feature interaction and global perception with efficient computational overhead.
 
     \item We propose a spatial-frequency collaborative enhancement mechanism. Specifically, the Fast Fourier Detail Perception module (FFDP) utilizes the Fast Fourier Transform (FFT) to map spatial-domain features into the frequency domain, thereby modeling global frequency information and enhancing the model's responsiveness to regions with grayscale variations. Meanwhile, the Haar Wavelet Transform Detail Enhancement Block (HWDE), based on the Haar wavelet transform, further captures high-frequency components related to edges and textures, reinforcing local detail representation. This strategy effectively integrates spatial texture with frequency-domain structural information, thereby enhancing the model's robustness and boundary perception capabilities.
\end{enumerate}

\begin{figure*}[htbp]
    \centering
    \includegraphics[width=\textwidth]{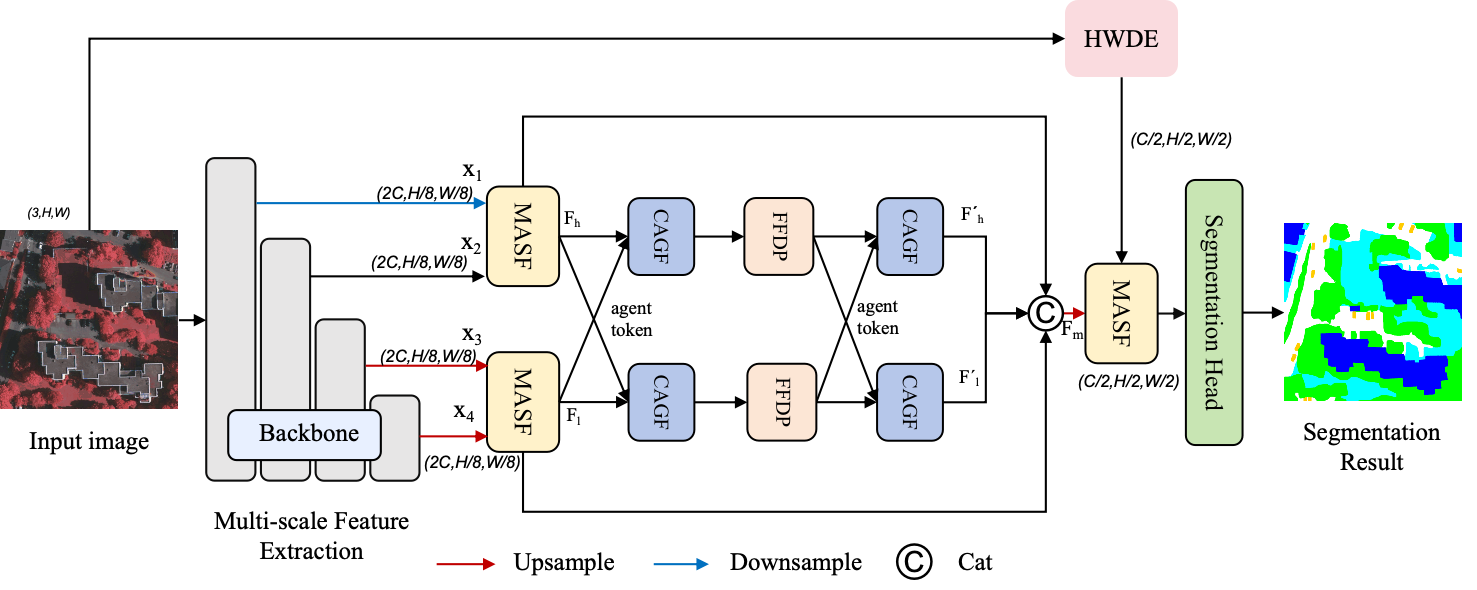}
    \caption{The overall network architecture of our proposed FSDENet. Specifically, using ConvNeXt-Small to extract multi-scale features, unifying the extracted features to a scale size of $X_2$, using MASF for receptive field alignment of features at different scales, using CAGF for global information supplementation and feature interactions, using FFDP to introduce frequency-domain information in the global information efficiently, and finally fusing it with information after detail enhancement via HWDE. The final segmentation result is generated by the segmentation head.}
    \label{fig:architecture}
\end{figure*}
\section{RELATED WORKS}
\subsection{CNN and Transformer Based Remote Sensing Image Semantic Segmentation}

Remote sensing images present unique challenges, such as complex backgrounds, small targets, and shadow interference. Traditional methods often struggle with these complexities due to their limited receptive fields, requiring not only semantic information but also rich details and global context. To address these challenges, various approaches have been explored, including expanding the receptive field \cite{espnet},\cite{cite_receptive_field1},\cite{cite_receptive_field2},\cite{cite_receptive_field3} and leveraging boundary information \cite{detail-enhanced},\cite{edge1},\cite{edge2}\cite{jstart_overview}. UNet \cite{Unet} incorporates skip connections to capture richer contextual information, making it a widely adopted segmentation framework. Furthermore, methods such as those proposed by Shi et al. \cite{PsPnet} and Chen et al. \cite{DeeplabV3++} utilize pyramid pooling to extract multi-scale image context, effectively aggregating local and global information across different feature scales.

CNN-based models primarily extract local features and initially lack a global understanding of the input image. However, with the introduction of Vision Transformer (ViT) \cite{ViT}, Transformer-based methods \cite{HIRI-VIT}, \cite{agent} have enabled models to capture global information from the outset. Several approaches integrate CNN and Transformer architectures, such as TransUNet \cite{Transunet} for medical image segmentation and UNetFormer \cite{Unetformer} for remote sensing image segmentation, which incorporate Transformer structures in the encoder and decoder, respectively. These methods effectively leverage both local and global information, demonstrating success in image segmentation tasks.

Currently, most remote sensing image segmentation methods utilize hybrid models that combine Transformers and CNNs \cite{jstart_cnn_transformer}\cite{jstart_swintransformer}, such as ConvLSR-Net \cite{LSRFormer} and CMTF-Net \cite{cmtfnet}. Additionally, some models adopt a pure Transformer architecture, including Segmenter \cite{Segmenter} and SwinUNet \cite{swinunet}. However, due to the high resolution of remote sensing images, self-attention mechanisms incur significant computational costs. Therefore, it is essential to develop methods with lower computational complexity to enhance training and inference efficiency. In this work, we address this challenge by employing an improved Cross-Agent Attention mechanism for global feature mapping in a single decoder stage.

\subsection{Haar Wavelet Transform and Fast Fourier Transform in Image Processing}
Haar Wavelet transform and fast Fourier transform are commonly used in signal processing, compression, and denoising tasks. A growing number of methods have been employed for image processing in recent years. For example, Tatsunami and Tak et al.\cite{FFT_base_tokenmix} designed an FFT-based Token Mixer to replace Multi-head Self-Attention (MHSA) and proved that their model has similar representations and properties to those using MHSA. Cui et al.\cite{okmi} proposed an omni-kernel that utilizes a combination of FFT and CNN modules to deal with image restoration tasks and achieve state-of-the-art (SOTA) results. Xu et al.\cite{HWD} designed HWD instead of downsampling to retain more detailed information using Haar Wavelet transform. In contrast, Finder et al. \cite{WaveConv} designed a backbone network based on the Haar Wavelet transform to obtain a larger receptive field.
\begin{figure}[t]
    \centering
    \includegraphics[width=\linewidth]{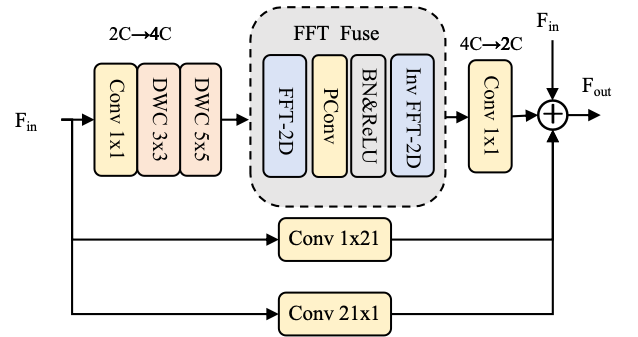}
    \caption{An illustration of the FFDP Block.}
    \label{fig:FFDP}
\end{figure}
In the task of remote sensing image segmentation, Yang et al. proposed SFFNet \cite{SFFnet}, which integrates information processed by the Haar Wavelet Transform with both local and global features, achieving significant improvements. However, SFFNet does not fully exploit the characteristics of transformed frequency domain information. The Haar Wavelet Transform is particularly effective for capturing local variations, such as edges, while the Fast Fourier Transform (FFT) excels at analyzing frequency components across the entire image.

In our work, we leverage both transforms to enhance segmentation accuracy. Specifically, we employ the Haar wavelet transform to refine image edge details and utilize the Fast Fourier Transform to enhance the model's ability to capture global grayscale variations. This dual-transform approach helps mitigate challenges in remote sensing imagery related to low contrast and blurred edge detection, particularly in scenarios with shadows or occlusions.

\section{METHODS}
As illustrated in \autoref{fig:architecture}, the proposed architecture of our network is derived from the UNet architecture, which has been observed to produce excessive redundant information. Inspired by SegFormer\cite{SegFormer}, we propose the following architecture.

\subsection{FSDENet Structure}
Specifically, given a high-resolution remote sensing image, we first partition it into a set of sub-images of size $3\times H \times W$, where three corresponds to the RGB channel. By performing sufficient spatial feature extraction using ConvNeXt\cite{ConvNext}, we obtain four multi-level outputs of different sizes: $x_1 \in \mathbb{R}^{ C\times(H/4) \times (W/4)}$, $x_2 \in \mathbb{R}^{2C \times (H/8) \times (W/8)}$, $x_3 \in \mathbb{R}^{4C \times (H/16) \times(W/16)}$, $x_4 \in \mathbb{R}^{8C \times (H/32) \times (W/32)}$and C = 96. In the decoder section, FSDENet does not adopt the UNet\cite{Unet} network structure because the UNet network contains too much redundant information. Inspired by  by the Segformer\cite{SegFormer}, we adjust the outputs at all levels to the size of $x_2$. Unify ($x_1,x_2$) and ($x_3,x_4$) by MASF to get $F_h,F_l$. 
\begin{equation}
\left\{
\begin{aligned}
    F_h = f_{masf}(x_1,x_2)\\
    F_l = f_{masf}(x_3,x_4)
\end{aligned}
\right.
\end{equation}
Here $f_{masf}(F_1,F_2)$ denotes the MASF block, $F_h,F_l \in \mathbb{R}^{2C\times(H/8)\times (W/8)}$.

\begin{figure}[t]
    \centering
    \includegraphics[width=\linewidth]{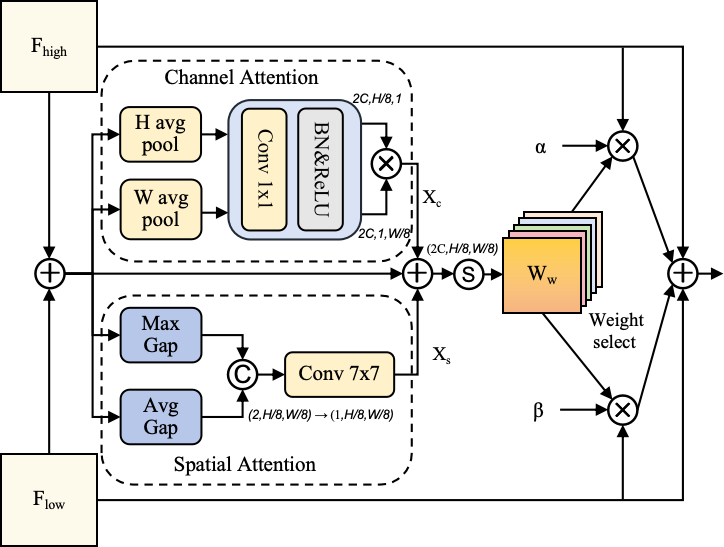}
    \caption{An illustration of the MASF Block}
    \label{fig:MASF}
\end{figure}

Global features are captured by two CAGF global mapping branches, leveraging the interaction between shallow features and deep features. This enables shallow information to possess deep semantic information, while deep features contain finer edge textures, thereby transforming raw data into more discriminative feature information. Subsequently, FFDP is employed to effectively introduce frequency-domain information, addressing the common issue of insufficient feature diversity in self-attention mechanisms. This enhances the network's segmentation accuracy in regions with intense grayscale variations.

\begin{equation}
\left\{
\begin{aligned}
    F_{hh} &= f_{ffdp}(f_{cagf}(F_h, F_l))\\
    F_{ll} &= f_{ffdp}(f_{cagf}(F_l, F_h))\\
    F_h' &= f_{cagf}(F_{hh}, F_{ll})\\
    F_l' &= f_{cagf}(F_{ll},F_{hh})
\end{aligned}
\right.
\end{equation}
where $f_{ffdp}(F_1)$ denotes the FFDP block, $f_{cagf}(F_2,F_3)$ denotes the CAFG block, and $F_3$ stands for the agent token.

Here, the feature merging is done directly using convolution, expressed as follows:
\begin{equation}
    F_m = \text{Cat}(F_h', F_l', F_h, F_l)
\end{equation}
Here  $F_m \in \mathbb{R}^{2C\times \frac{H}{8} \times \frac{W}{8} }$.

Finally, the extracted interactive global features are merged with the HWDE frequency domain detail-enhanced feature $Y$, which is then sampled to the original image size by the segmentation head to obtain the final segmentation result.
\begin{equation}
    Y = (\text{Cat}(f_{hwde}(x), F_m))
\end{equation}

\subsection{Mulit-Attention Select Fusion Block (MASF)}

\begin{figure*}[htbp]
    \centering
    \includegraphics[width=\textwidth]{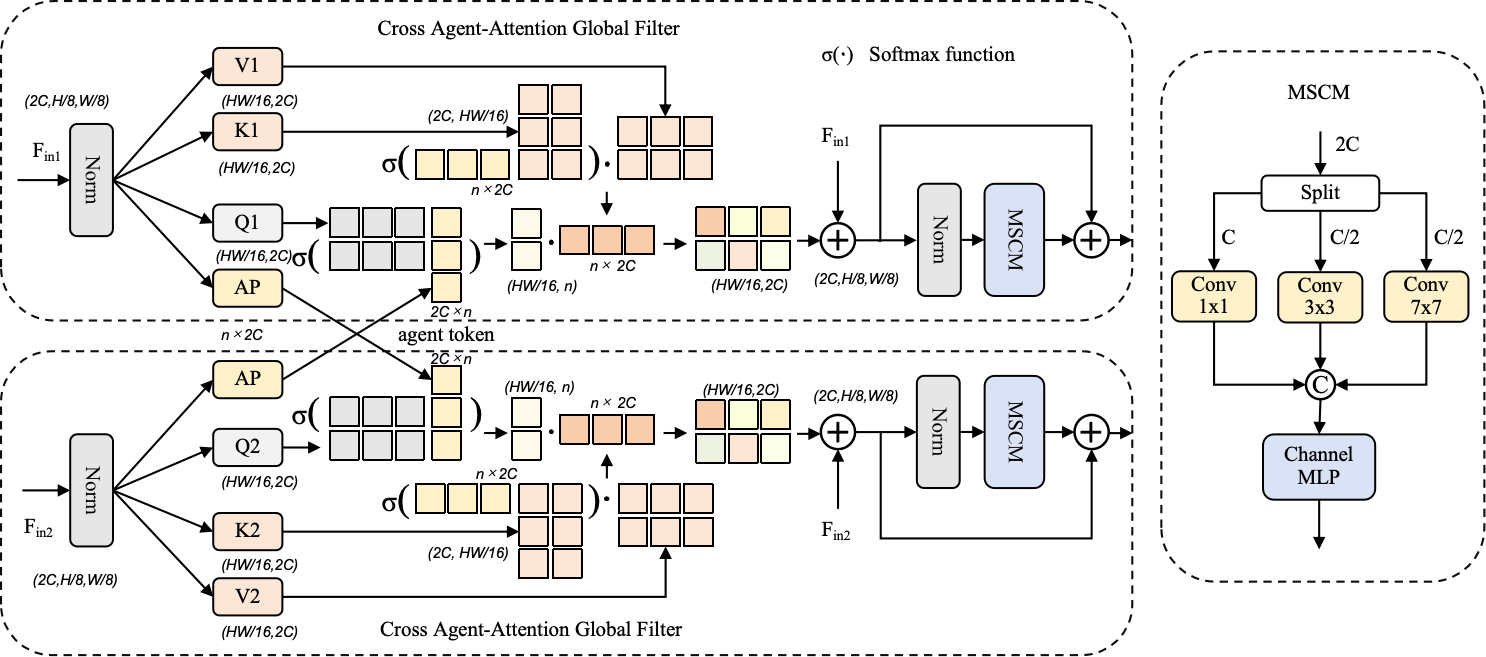}
    \caption{An illustration of the CAGF Block.}
    \label{fig:CAGF}
\end{figure*}
We adopt an encoder-decoder-like architecture. It is observed that fusing multi-scale features extracted by the encoder is critical for improving the model’s ability to recognize objects of varying sizes. Shallow features typically retain rich edge and texture details, while deeper features encode more abstract semantic representations. However, as the network deepens, the influence of shallow features diminishes, leading to significant degradation in the representation of small objects and boundary regions. Moreover, simple operations such as element-wise addition, concatenation, or naive mixing fail to address the inherent mismatch between features before fusion.

To effectively mitigate the semantic masking effect in deep-shallow feature fusion, we propose a Multi-scale Adaptive Selection Fusion (MASF) module. This module optimizes features through spatial and channel attention mechanisms, weighting spatial attention through channel attention to assign distinct importance weights to each channel. It incorporates prior information and achieves effective feature fusion using learnable attention maps and a gating mechanism.

% As shown in \autoref{fig:MASF}, MASF operates on input features $F_{high}, F_{low} \in \mathbb{R}^{C \times H \times W}$ denotes the proceeding input features and fused them by a 1x1 convolution:
As shown in \autoref{fig:MASF}, MASF operates on input features $F_{high}, F_{low} \in \mathbb{R}^{2C \times H/8 \times W/8}$ denotes the preceding input features and fused them by a 1x1 convolution:
\begin{equation}
    F_{in} = \delta_{1 \times 1}(F_{high} + F_{low}) 
\end{equation}
For ease of description, we define:
\begin{equation}
    \varphi(x) = \text{relu}(\text{bn}(\delta_{1 \times 1}(x)))
\end{equation}
For channel attention, we use pooled kernels of size (H, 1) and (1, W) to encode each channel along the horizontal and vertical coordinates, extracting the essential features in the entire feature mapping for each channel in the H and W directions:
\begin{equation}
    X_c = \varphi(X_{HAP}^c) \cdot \varphi(X_{WAP}^c)
\end{equation}
Here, $X_{HAP}^c$ and $X_{WAP}^c$ represent features obtained following a global average pooling operation across channel dimensions in the H and W directions.

For spatial attention, we pool the dimensions of the channel using maximum pooling and average pooling operations, and use 7x7 large kernel convolution to enhance local correlation between spatial features:
\begin{equation}
    X_s = \delta_{7 \times 7}(Cat(X_{GAP}^s, X_{GMP}^s))
\end{equation}
Here, $\delta_{k \times k}(\cdot)$ denotes the convolution with a kernel size of k × k;  $X_{GAP}^s $ and $ X_{GMP}^s$ represent features obtained following a global average pooling operation across channel dimensions, a global maximum pooling operation across special dimensions. 

Then, the spatial attention weights are spliced with the inputs in the channel direction and the prior knowledge is introduced to obtain the weights $W_w$ by Sigmoid function. 
\begin{equation}
    W_w = Sigmod(X_c + X_s + F_{in})
\end{equation}
Finally, the feature fusion is performed by two trainable parameters with the following equation:
\begin{equation}
    F_{fuse} = \delta_{1 \times 1}(F_{low} \cdot \alpha \cdot W_w + F_{high} \cdot \beta \cdot W_w + F_{low} + F_{high})
\end{equation}
Here, use a 1x1 convolution to adjust the correlation between the feature channels; $\alpha$ and $\beta$ are trainable parameters.

\begin{figure*}[htbp]
    \centering
    \includegraphics[width=\textwidth]{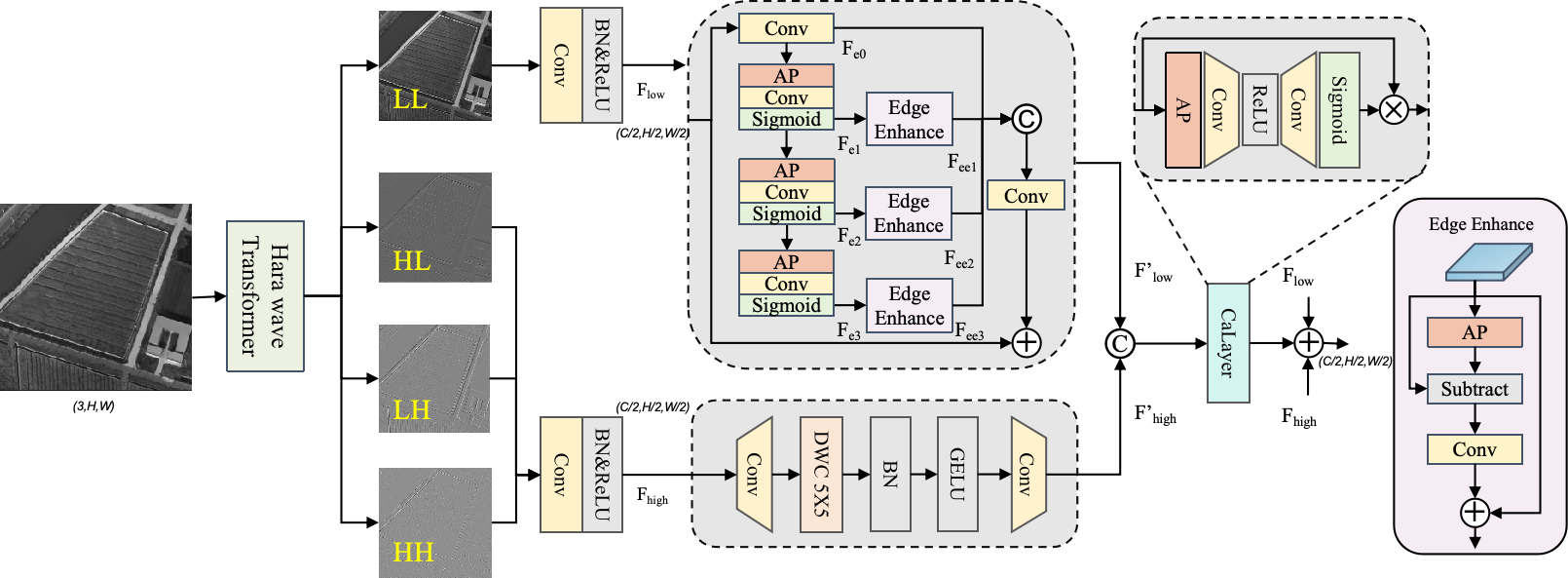}
    \caption{An illustration of the HWDE Block.}
    \label{fig:HWDE}
\end{figure*}

\subsection{Cross Agent-Attention Global Filter (CAGF): }
% To reduce the quadratic complexity introduced by self-attention mechanisms, we employ agent tokens to lower computational complexity to linear scale while effectively extracting global information. This approach achieves efficient long-range dependency modeling by compressing interaction patterns through learnable proxy tokens, maintaining the model's ability to capture comprehensive contextual relationships without sacrificing representational capacity.
To address the challenge of capturing long-range dependencies in high-resolution remote sensing images while mitigating the prohibitive computational cost of standard self-attention mechanisms, we introduce the Cross Agent-Attention Global Filter (CAGF). Remote sensing scenes often contain large-scale objects, complex spatial structures, and diverse contextual relationships that demand effective global context modeling. However, traditional self-attention suffers from quadratic complexity concerning input size, making it impractical for processing large-scale feature maps.

To overcome this, CAGF adopts a learnable agent token mechanism that compresses spatial interaction patterns into a set of representative proxy tokens, significantly reducing computational complexity from quadratic to linear scale. This not only ensures computational efficiency but also enables effective feature interaction between shallow and deep layers. By exchanging and aggregating global semantic cues through these agent tokens, CAGF preserves the model’s ability to capture comprehensive contextual information and enhances semantic consistency across scales—particularly beneficial in remote sensing scenes with extensive spatial variability and class imbalance.

As illustrated in \autoref{fig:CAGF}. Here 49×49 average pooling is used to generate agent tokens from input features $F_h  \in \mathbb{R}^{2C \times H/8 \times W/8}$ and $F_l  \in \mathbb{R}^{2C \times H/8 \times W/8}$, which can be represented as:
\begin{equation}
    A_h,A_l= AP_{49\times49}(F_h), AP_{49\times49}(F_l)
\end{equation}
where $ A_h,A_l\in \mathbb{R}^{n \times 2C}$ is our newly defined agent tokens; $AP_{k\times k}(\cdot)$ denotes k×k average pooling.

In the attention computation process, the agent tokens from the two features are exchanged, with q and k being used for Softmax attention computation:
\begin{equation}
    O_h = \sigma(Q_h A_l^T) \ \sigma(A_l K_h^T) \, V_h,
    O_l = \sigma(Q_l A_h^T) \ \sigma(A_h K_l^T) \, V_l
\end{equation}
Where $\sigma(\cdot)$ denotes Softmax function; $ Q,K,V\in \mathbb{R}^{N \times 2C}$ denote query, key and value matrices.

In Multi-scale Channel MLP(MSCM), to complement the feature diversity that linear attention lacks, we classify the features into $ F_1 \in \mathbb{R}^{(2C/2) \times H/8 \times W/8}$, $ F_2, F_3 \in \mathbb{R}^{(2C/4) \times H/8 \times W/8}$ by channel dimensions before Channel MLP and split them using 1x1 convolution, 5x5 convolution, 7x7 convolution to increase the feature diversity, the formula is as follows:
\begin{equation}
    F = CAT(\delta_{1 \times 1}(F_1),\delta_{5 \times 5}(F_2),\delta_{7 \times 7}(F_3))
\end{equation}
The final MLP section uses the Channel MLP\cite{li2023moganet}.

\subsection{Fast Fourier Detail Perception Block (FFDP)} 
Traditional attention mechanisms and convolutional neural networks  primarily model contextual information in the spatial domain, which limits their ability to effectively capture large-scale grayscale variations. This issue becomes particularly prominent in shadowed or low-contrast regions, where models often exhibit instability and struggle to detect gradual grayscale transitions or non-local texture patterns across regions. In contrast, the frequency domain is inherently sensitive to intensity variations and periodic texture structures, making it especially suitable for capturing global grayscale patterns and structural textures that span across spatial regions. Integrating frequency-domain features can not only compensate for the locality limitation of spatial-domain modeling but also enhance the model’s holistic understanding of structural details.

We innovatively integrate the Fast Fourier Transform into the FFDP module to overcome the limitations of traditional spatial-domain processing by employing a frequency-domain analysis method. This mathematical transform enables the accurate mapping of signals from spatial representations to frequency domain components, allowing the model to capture both high-frequency edge features and low-frequency texture information of images in parallel.

As illustrated in \autoref{fig:FFDP}. First, Li et al. demonstrated that serially applying two small convolutional kernels can achieve a receptive field equivalent to that of a larger kernel, while requiring less computational cost \cite{LSKNet}. Therefore, we use two serially connected 3×3 and 5×5 depth-wise convolutions to enhance the receptive field at a relatively low cost. Let the input feature $F_{in} \in \mathbb{R}^{2C \times (H/8)\times (W/8)}$, we have:
\begin{equation}
    F_1 = \psi_{5 \times 5}(\psi_{3 \times 3}(\delta_{1 \times 1}(F_{in} )))
\end{equation}
Where $\psi_{k \times k}(\cdot)$ is depth-wise convolution with a kernel size of k × k. $F_1 \in \mathbb{R}^{4C \times (H/8)\times (W/8)}$. 

Recalling the Discrete Fourier Transform (DFT), given a feature map $X \in \mathbb{R}^{C \times H\times W}$, DFT can be formalized as follows:
\begin{equation}
\mathcal{F}(u, v) = \frac{1}{\sqrt{HW}} \sum_{h=0}^{H-1} \sum_{w=0}^{W-1} X(h, w) e^{-j2\pi \left( \frac{hu}{H} + \frac{wv}{W} \right)}
\end{equation}
where $\mathcal{F}(u, v)$ is based on Fourier space as the complex component; u and v are
the coordinates of Fourier space.

In order to achieve learnable feature modulation in the frequency domain space, we choose the lightweight partial convolution kernel for frequency domain feature modulation, use the BN layer to achieve modal normalization of complex features, and the ReLU function acts on the magnitude spectrum of the normalized features to achieve adaptive thresholding of the frequency domain feature filtering with the following specific formula:
\begin{equation}
    F_2 = \delta_{1 \times 1}(IFFT(relu(bn(Pconv(FFT(F_1))))))
\end{equation}
Here, FFT and IFFT refer to fast Fourier transforms and their inverse operations. Pconv is Partial Convolution, where using a convolutional kernel size of 1x1 for inter-channel interaction reduces the number of channels from $2C$ to $C$. 

To meet the feature extraction demands for elongated targets like roads and rivers, we employ two (1×21) and (21×1) strip-shaped depthwise separable convolutions. These operations respectively expand receptive fields along the horizontal and vertical axes of slender targets, effectively capturing continuous features in shadowed or occluded regions while enhancing contextual relationships among internal pixels of elongated objects without significantly increasing computational costs. Finally, the processed frequency-domain features are fused with spatial-domain features to generate an output feature map containing rich global and edge texture information. The specific formula is as follows
\begin{equation}
    F_{out} = \delta_{1 \times 1}(F_2 + \delta_{1 \times 21}(F_{in}) + \delta_{21 \times 1}(F_{in})+F_{in})
\end{equation}
$F_{out} \in \mathbb{R}^{2C \times (H/8)\times (W/8)}$ is the feature fusion output.

\subsection{Haar Wavelet transform Detail Enhancement Block (HWDE)}
% To enhance edge segmentation accuracy, we propose a frequency-aware feature enhancement framework based on wavelet decomposition. This approach utilizes the Haar wavelet transform to explicitly separate visual information into low-frequency structural components and high-frequency edge details, allowing for specialized processing within each frequency domain.

In remote sensing imagery, challenges such as shadows, occlusions, and blurred object boundaries are prevalent, particularly around the edge regions of targets. These critical structural cues are typically represented by high-frequency components in the image. However, conventional convolutional operations, especially under multiple downsampling layers, tend to attenuate or even discard high-frequency information, which adversely affects the model’s ability to accurately localize object boundaries and detect small-scale targets. To address this limitation, this paper introduces a frequency decomposition mechanism based on the Haar wavelet transform, which separates the original spatial features into low-frequency components that capture global structural information and high-frequency components that emphasize edges and textures. By selectively enhancing and reconstructing these frequency-specific features, the model is able to more effectively recover boundary details and improve its sensitivity to object contours, edge transitions, and small-scale features, thereby enhancing the overall accuracy and robustness of semantic segmentation.

As illustrated in\autoref{fig:HWDE}. Haar wavelet transform can decompose the image into low-frequency feature LL, high-frequency horizontal feature HL, vertical feature LH, and diagonal feature HH:
\begin{equation}
    [LL,HL,LH,HH] = HWT(X)
\end{equation}
\begin{equation}
\left\{
\begin{aligned}
    F_{low} &= relu(bn(\delta_{1 \times 1}(LL)))\\
    F_{high} &= relu(bn(\delta_{1 \times 1}(HL+LH+HH)))
\end{aligned}
\right.
\end{equation}
Here $HWT(\cdot)$ denotes Haar wavelet transform, $x \in \mathbb{R}^{3 \times H\times W}$ is the input feature. $LL,HL,LH,HH \in \mathbb{R}^{C \times (H/2)\times (W/2)}$

Low-frequency information(LL) usually represents the overall structure of an image and is primarily used to recover large-scale features, such as contours and backgrounds. In this context, low-frequency information plays a pivotal role. To address the missing image detail and edge features and enhance the model's ability to capture object boundaries, we employ average pooling and convolution to extract multi-scale edge information from the Low-frequency information. 
\begin{equation}
\left\{
\begin{aligned}
    F_{e0} &= \delta_{3 \times 3}(LL)\\
    F_{ei} &= Sigmoid(\delta_{1 \times 1}(AP_{3 \times 3}(F_{e(i-1)}))), i=1,2,3
\end{aligned}
\right.
\end{equation}
Edge perception is further refined at each scale by an edge enhancer, which emphasizes the critical boundaries of the object. Which can be represented as follows:
\begin{equation}
    f_{ee}(x) = x + \delta_{1 \times 1}(x - AP_{3 \times 3}(x))\\
\end{equation}
\begin{equation}
    F_{eei} = f_{ee}(F_{ei}),i=1,2,3
\end{equation}
where $AP_{k\times k}$ denotes k×k average pooling, $F_{eei}$ is the feature after detail enhancement. 
\begin{table}[ht]
    \centering
    \renewcommand{\arraystretch}{1.2} 
    \setlength{\tabcolsep}{15pt}
    \captionsetup{font={small}}
    \caption{COMPARISON WITH THE SOTA METHODS ON THE ISAID DATASET}
    \begin{tabular}{ccc}
        \toprule
        \textbf{Method} & \textbf{Backbone} & \textbf{mIoU(\%)} \\
        \midrule
        HRnet\cite{HRNet} & HRnet-32 & 62.3 \\
        DeeplabV3+\cite{DeeplabV3++} & ResNet50 & 61.2 \\
        SFNet\cite{SFNet} & ResNet50 & 64.3 \\
        VB+R-UperNet\cite{wang2022empirical} & ViTAE-B & 64.5 \\
        PFNet\cite{PFNet} & ResNet50 & 64.3 \\
        SegFormer\cite{SegFormer} & MiT-B4 & 67.2 \\
        SegNeXt-L\cite{Segnext} & MSCAN-L & 70.3 \\
        FarSeg++\cite{Farseg++} & MiT-B2 & 67.9 \\
        RssFormer\cite{Rssformer} & HRnet-32 & 65.9 \\
        \hline
        FSDENet & ConvNeXt-Small & 70.3 \\
        \bottomrule
    \end{tabular}

    \label{tab:isaid}
    
\end{table}
\begin{figure}[b]
    \centering
    \includegraphics[width=\linewidth]{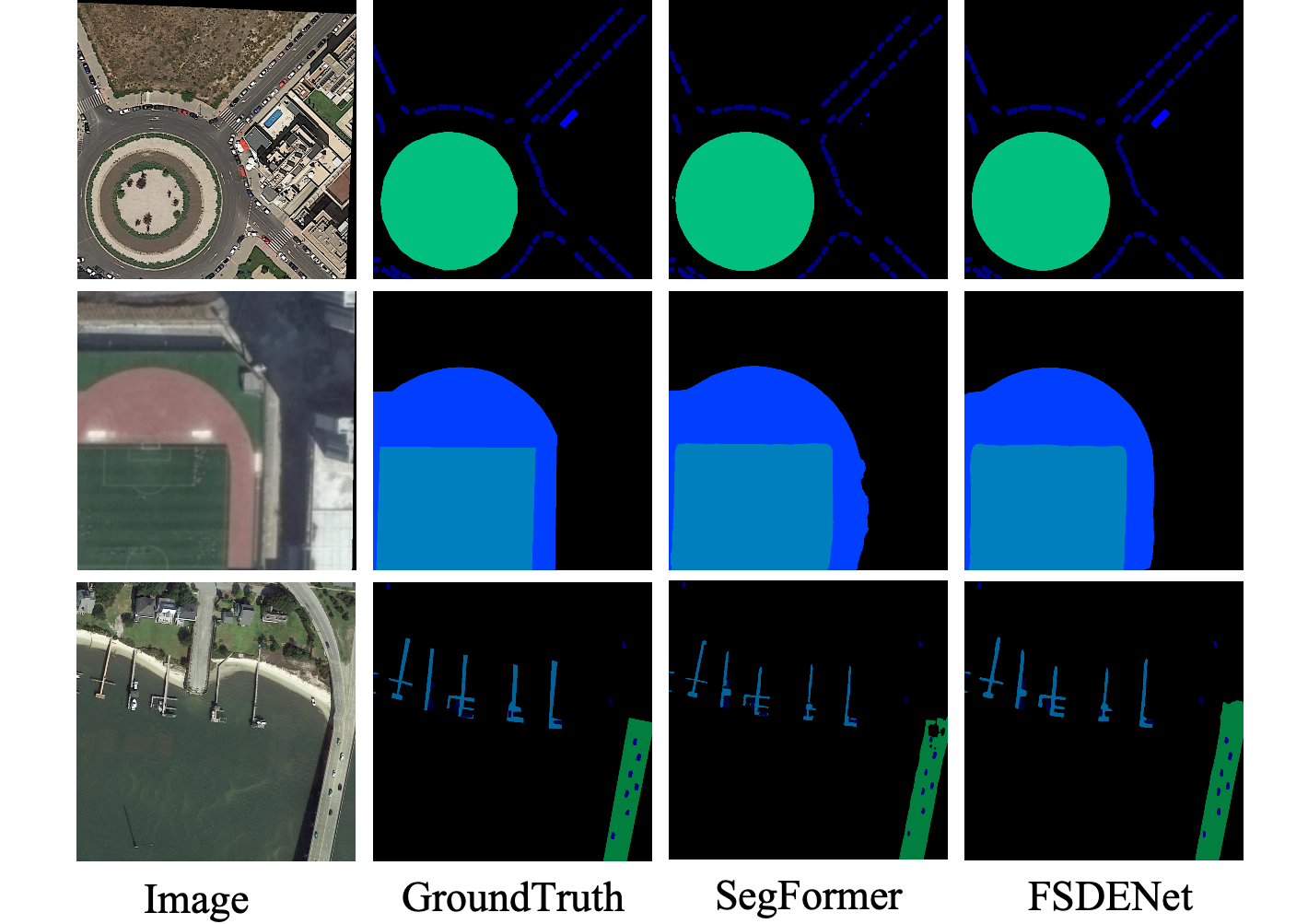}
    \captionsetup{font={small}}
    \caption{Qualitative comparisons between ours and SegFormer on the iSAID dataset}
    \label{fig:isaid}
\end{figure} 
The extracted multi-scale edge information is fused with the features of the main branch to enhance the fineness of the low-frequency features:
\begin{equation}
    F'_{low} = F_{low} + \delta_{1 \times 1}(CAT(F_{e0},F_{ee1},F_{ee2},F_{ee3}))
\end{equation}

High-frequency information(HL, LH, HH) usually reflects an image's local details. Through $5 \times
 5$ depth-wise convolution and reverse bottleneck design, these high-frequency features are efficiently extracted while
being lightweight to enhance the details of high-frequency information:
\begin{equation}
    F'_{high} = \delta_{1 \times 1}(glue(bn(\psi_{5 \times 5}(\delta_{1 \times 1}(F_{high})))))
\end{equation}
Here, the first convolution increases the channels from $C$ to $2C$, and the final convolution changes the number of channels back to $C$.

The high and low-frequency information($F'_{high}$,$F'_{low}$) is fused by Calayer, and the final extracted multi-scale edge information is fused with the features of the main branch, which finally improves the fineness of the features:
\begin{equation}
\left\{
\begin{aligned}
    f_{CaLayer}(x) = x \cdot Sigmoid(\delta_{1 \times 1}(relu(\delta_{1 \times 1}(AP(x))))) \\
    F_{out} = f_{CaLayer}(F'_{low} + F'_{high}) + F_{low} + F_{high}
\end{aligned}
\right.
\end{equation}
Here $F_{out} \in \mathbb{R}^{C\times (H/2)\times (W/2)}$ denotes the feature that has undergone detailed feature enhancement using the Haar Wavelet transform.

\section{DATASETS AND EXPERIMENT SETTINGS}
\subsection{Datasets}
1) \textbf{LoveDA}\cite{LoveDA}: The LoveDA dataset was constructed using high-resolution 0.3 m images acquired in July 2016 from the cities of Nanjing, Changzhou, and Wuhan. Each image has a resolution of 1024 × 1024 pixels with no overlap. The dataset comprises seven land cover categories: buildings, roads, water, barren land, forest, agriculture, and background. It includes 18 complex urban and rural scenes, containing a total of 166,768 annotated objects. We divided the dataset into 2,522 training images, 1,669 validation images, and 1,796 test images.

2) \textbf{Vaihingen}: The Vaihingen dataset comprises 33 highly detailed spatial resolution TOP image tiles, each with an average size of 24.94 × 2064 pixels. The dataset includes five foreground classes (impervious surfaces, buildings, low vegetation, trees, cars) and one background class (clutter). In our experiments, we exclusively used the TOP image tiles. The experiments were conducted using the following IDs: 2, 4, 6, 8, 10, 12, 14, 16, 20, 22, 24, 27, 29, 31, 33, 35, and 38, with ID 30 used for validation and the remaining 15 images designated for training. The image tiles were cropped into smaller patches of 1024 × 1024 pixels to facilitate processing.

3) \textbf{Potsdam}: The Potsdam dataset consists of 38 TOP image blocks with a very high spatial resolution and an image size of 6000x6000 pixels. This dataset covers the same category information as the Vaihingen dataset. We selected image blocks with the IDs 2\_13, 2\_14, 3\_13, 3\_14, 4\_13, 4\_14, 4\_15, 5\_13, 5\_14, 5\_15, 6\_13, 6\_14, 6\_15, 7\_13 for testing and validating them using the image block with the ID 2\_10. The remaining 22 image blocks (excluding the incorrectly annotated 7\_10 image blocks) were used for training. During the experiments, only the red, green, and blue spectral bands were used, and the original image blocks were cropped to a size of 1024×1024 pixels.

4) \textbf{ISAID}: The iSAID dataset contains high-resolution remote sensing images from different geographical regions and covers many complex scenarios and diverse target distributions. The dataset includes 2,806 remotely sensed images with more than 650,000 labeled instances. The resolution of the images ranges from 800x800 to 4000x4000. Following the experimental setup\cite{LSRFormer}, the dataset is divided into 1411/458/937 images for train/val/test. Each image is overlapped and segmented into sub-images of size 896 × 896 with a step size of 512 by 512.

\subsection{Implementation Details}
Following the previous work\cite{Unetformer}, We used the AdamW algorithm with a cosine learning rate variation strategy for the optimizer, with a base learning rate of 6e-4. We trained our model on two NVIDIA Tesla V100 16G graphics cards. For the Vaihinge and Potsdam datasets, the images were randomly cropped into small blocks of 512 × 512, and the training epoch was set to 105. The training epochs for the ISAID and LoveDA datasets were 60 and 30, respectively (with LoveDA also randomly cropped into 512 × 512 chunks and ISAID used at its original size). Enhancement techniques such as random scaling ([0.5,0.75,1.0,1.25,1.5]), random vertical flip, random horizontal flip, and random rotation were used during the training process, and the batch size was set to 8 (the batch size for ISAID was set to 2).

\subsection{Evaluation Metrics}

Following previous work, we adopt the mean intersection over union (mIoU) as the primary evaluation metric for the iSAID and LoveDA datasets. For the Vaihingen dataset, we use mIoU, overall accuracy (OA), and mean F1 score (mF1) as evaluation indicators. The definitions of OA, mF1, and mIoU are as follows:

\begin{equation}
    \text{OA} = \frac{\text{TP} + \text{TN}}{\text{TP} + \text{FP} + \text{TN} + \text{FN}}
\end{equation}

\begin{equation}
    \text{mF1} = \frac{1}{k+1} \sum_{i=0}^{k} \frac{2\text{TP}}{2\text{TP} + \text{FP} + \text{FN}}
\end{equation}

\begin{equation}
    \text{mIoU} = \frac{1}{k+1} \sum_{i=0}^{k} \frac{\text{TP}}{\text{FN} + \text{FP} + \text{TP}}
\end{equation}
TP(true positives), FP(false positives), FN(false negatives), TN(true negatives). OA is the ratio of correctly predicted pixels to the total number of pixels.

\section{EXPERIMENTAL RESULTS AND ANALYSIS}
\subsection{Ablation Experiment on Modules}
To fully assess the performance of each component in the FSDENet model, we conducted a comprehensive series of ablation experiments. These experiments aimed to observe the effect of removing or adding individual components on the overall performance. To ensure the reliability and validity of the experimental results, we selected two widely used datasets, Vaihingen and Potsdam, for validation. In performing the ablation experiments, we focused on two key performance metrics: mIoU and mF1.
 \begin{table}[H]
    \centering
    \renewcommand{\arraystretch}{1.4} 
    \setlength{\tabcolsep}{11pt}
    \captionsetup{font={small}}
    \caption{COMPARISON OF DIFFERENT METHODS IN TERMS OF PARAMETERS, FLOPS AND FPS.}
    \begin{tabular}{cccc}
    \toprule
    \textbf{\textbf{Method}} & \textbf{Params(M)} & \textbf{FLOPs(G)} & \textbf{FPS} \\ \midrule
    DeeplabV3+~\cite{DeeplabV3++} & 59.3 & 260.6 & 32  \\
    ST-UNet~\cite{stunet} & 161.0    & --    & 7        \\
    SwinUNet~\cite{Swin} & 84.0 & 97.7 & 29     \\
    TransUNet~\cite{Transunet} & 105.9  & 168.9  & 27          \\
    Segformer~\cite{SegFormer} & 84.6 & 110.2 & 20          \\
    FT-UNetFormer~\cite{Unetformer}& 96.0 & 128.4  & 37          \\
    ConvSLR-Net ~\cite{LSRFormer} & 68.1 & \textbf{71.1}  & \textbf{46} \\
    \hline
    FSDENet & \textbf{58.51} & 87.57 & 35 \\ \bottomrule
    \end{tabular}
    \label{table:parameters_flops}
\end{table}
\begin{table}[H]
    \centering
    \captionsetup{font={small}}
    \renewcommand{\arraystretch}{1.4} 
    \setlength{\tabcolsep}{8pt}
    \caption{RESULTS OF ADDING INDIVIDUAL MODULES ON THE BASELINE MODEL ON THE VAIHINGEN DATASET}
    \begin{tabular}{@{}l|llll@{}}
    \toprule
    \textbf{Method} & \textbf{Params(M)} & \textbf{Flops(G)} & \textbf{mF1(\%)} & \textbf{mIoU(\%)} \\ \midrule
    Baseline & 50.53 & 49.3 & 91.08 & 83.82 \\
    Baseline+FFDP & 54.02 & 62.45 & 91.34 & 84.27 \\
    Baseline+MASF & 51.43 & 50.77 & 91.27 & 84.15 \\
    Baseline+HWDE & 50.65 & 56.73 & 91.29 & 84.19 \\
    Baseline+CAGF & 55.99 & 73.48 & 91.43 & 84.42 \\ \bottomrule
    \end{tabular}
    \label{table:baseline}
\end{table}
\begin{table}[H]
    \centering
    \renewcommand{\arraystretch}{1.4} 
    \setlength{\tabcolsep}{8pt}
    \captionsetup{font={small}}
    \caption{RESULTS OF SFFNET WITH INDIVIDUAL COMPONENTS REMOVED}
    \begin{tabular}{@{}l|ll|ll@{}}
    \toprule
    \multirow{2}{*}{\textbf{Method}} & \multicolumn{2}{c}{\textbf{Vaihingen}} & \multicolumn{2}{c}{\textbf{Potsdam}} \\
    \cmidrule(lr){2-3} \cmidrule(lr){4-5}
    & \textbf{F1(\%)} & \textbf{mIoU(\%)} & \textbf{mF1(\%)} & \textbf{mIoU(\%)} \\ \midrule
    FSDENet          & 91.61 & 84.71 & 93.35 & 87.73 \\
    FSDENet w/o FFDP & 91.46 & 84.47 & 93.2 & 87.48 \\
    FSDENet w/o MASF & 91.51 & 84.54 & 93.28 & 87.6 \\
    FSDENet w/o HWDE & 91.45 & 84.45 & 93.2 & 87.44 \\
    FSDENet w/o CAGF & 91.44 & 84.43 & 93.25 & 87.57 \\ \bottomrule
    \end{tabular} 
    \label{table:fsdenet}
\end{table}
\begin{figure}[H]
    \centering
    \includegraphics[width=\linewidth]{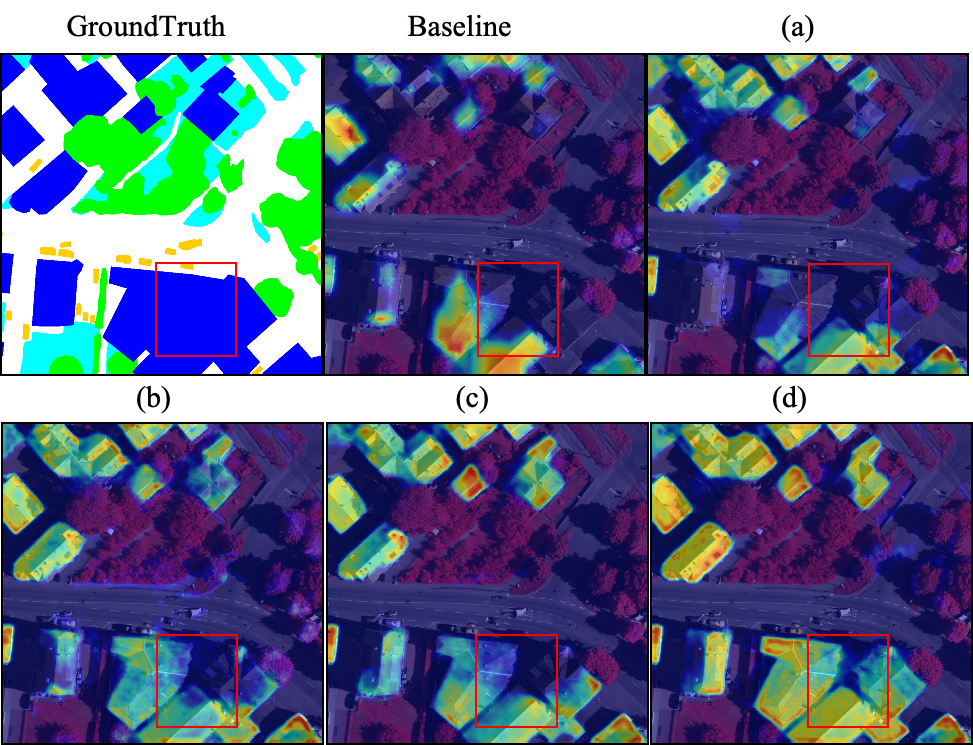}
    \captionsetup{font={small}}
    \caption{Grad-CAM visualization results for the “Building” class. (a)–(d) show the results by progressively adding CAGF, FFDP, MASF, and HWDE, with (d) representing the full model.}
    \label{fig:gam}
\end{figure}
\begin{table}[H]
    \centering
    \renewcommand{\arraystretch}{1.6} 
    \setlength{\tabcolsep}{8pt}
    \captionsetup{font={small}}
    \caption{RESULTS OF FSDENET WITH INDIVIDUAL COMPONENTS REMOVED}
    \begin{tabular}{cccccc}
    \toprule
    \multicolumn{4}{c}{Components} & \textbf{Vaihingen} & \textbf{Potsdam} \\
    CAGF & FFDP & HWDE & MASF & mIoU(\%) & mIoU(\%) \\
    \midrule
     &   &   &   & 83.82 & 86.82 \\
    \checkmark &   &   &   & 84.28 & 87.27 \\
    \checkmark & \checkmark &   &   & 84.41 & 87.45 \\
    \checkmark & \checkmark & \checkmark &   & 84.54 & 87.60 \\
    \checkmark & \checkmark &   & \checkmark & 84.44 & 87.44 \\
    \checkmark & \checkmark & \checkmark & \checkmark & 84.71 & 87.73 \\
    \bottomrule
    \end{tabular}
    \label{table:t}
\end{table}
\begin{table}[H]
    \centering
    \renewcommand{\arraystretch}{1.8} 
    \setlength{\tabcolsep}{3pt}
    \captionsetup{font={small}}
    \caption{ABLATION COMPARISON BETWEEN CAGF AND OTHER GLOBAL INFORMATION EXTRACTION MODULES IN TERMS OF PARAMETER COUNT, FLOPS, MIOU, AND F1 SCORES ON THE VAIHINGEN DATASET}
    \begin{tabular}{ccc|cc}
    \toprule
    \textbf{\textbf{Method}} & \textbf{mIoU(\%)}$\uparrow$ &\textbf{F1(\%)}$\uparrow$ & \textbf{Params(M)}$\downarrow$ & \textbf{FLOPs(G)}$\downarrow$  \\ \midrule
    Vit-Block\cite{ViT} & 84.44 &91.45&  3.56 & 3.85\\  
    Swin-Block\cite{Swin} & 84.52 &91.5& 0.445 &  7.76\\
    LSRFormer-Block\cite{LSRFormer} & 84.2&  91.31&0.372  &  1.22\\ 
    CAGF(Ours) & 84.71 &91.61& 0.333 & 5.47\\  
    \bottomrule
    \end{tabular}
    \label{table:parameters_flops}
\end{table}
\begin{figure}[H]
    \centering
    \includegraphics[width=\linewidth]{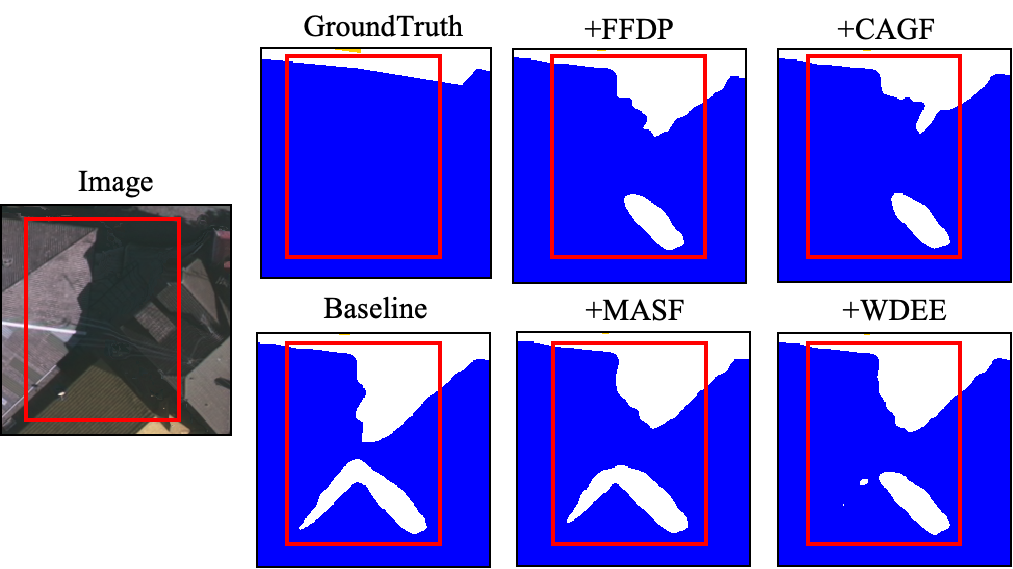}
    \captionsetup{font={small}}
    \caption{Local enlarged segmentation results after adding various components in Baseline.}
    \label{fig:baseline}
\end{figure}
\begin{figure}[H]
    \centering
    \includegraphics[width=\linewidth]{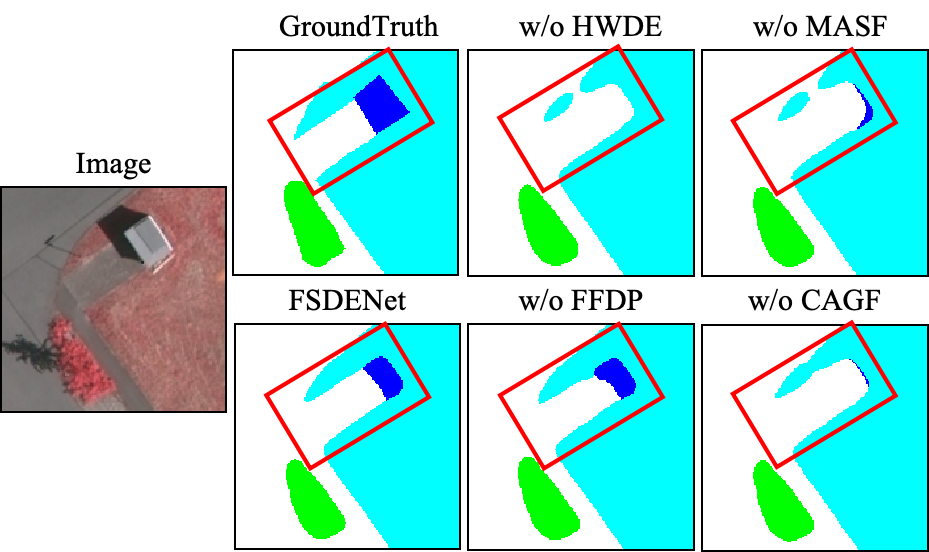}
    \captionsetup{font={small}}
    \caption{Local enlarged segmentation results of removing various components in FSDENet.}
    \label{fig:FSDENet}
\end{figure}
\autoref{table:baseline} shows the performance changes following adding a single module to the baseline model, while \autoref{table:fsdenet} reflects the effect of removing a single module from the FSDENet model. With the addition of the CAFG module to the benchmark model, the mF1 and mIoU metrics achieve an improvement of 0.35\% and 0.6\%, respectively, while the number of parameters and computational complexity increase by 5.45M and 24.18G, respectively. This is closely followed by the effect of adding the FFDP module. This phenomenon can be attributed to the fact that the benchmark model relies solely on the convolution operation to extract multi-scale features, which limits its ability to model only local information. In contrast, the FFDP and CAFG modules can add global information to the model.

On the Vaihingen dataset, removing the CAGF module results in a more significant decrease in key performance metrics than removing the FFDP module. However, on the Potsdam dataset, where the image resolution is higher, the situation is different. It is worth noting that on the Potsdam dataset, removing the HDEE module results in more performance degradation than removing the CAGF module. This may be because the Vaihingen dataset has a relatively simple sample distribution, whereas the Potsdam dataset has more complex foreground and background distributions. On such a dataset, the advantage of the frequency domain information carried by the FFDP and HWDE modules in processing texture details becomes apparent As shown in \autoref{fig:gam}, the CAGF module significantly enhances global semantic consistency by improving the model's holistic attention to the building class after incorporating global contextual information. Meanwhile, the MASF module facilitates better detail recovery, particularly along object boundaries. Furthermore, the FFDP and HWDE modules demonstrate stronger responses to shadowed and low-contrast regions through frequency-domain enhancement, effectively mitigating segmentation errors caused by weak grayscale variations.
\begin{table*}[htbp]
	\centering
    \captionsetup{font={small}}
	\caption{COMPARISON WITH THE SOTA METHODS ON THE LOVEDA DATASET}
	\renewcommand{\arraystretch}{1.4} 
    \setlength{\tabcolsep}{9.5pt}
	\begin{tabular}{c|c|ccccccc}         \toprule
		\multirow{2}{*}{\textbf{Method}} & \multirow{2}{*}{\textbf{mIoU} } & \multicolumn{6}{c}{\textbf{IoU(\%)}}                                                 \\ 
		\cline{3-9}& & \textbf{Background} & \textbf{Building} & \textbf{Road}  & \textbf{Water} & \textbf{Barren} & \textbf{Forest} & \textbf{Agricultural}  \\ 
		\cline{1-3}\cline{4-9}
        Fcn\cite{Fcn} & 45.56 & 51.19 & 50.78 & 47.65 & 57.66 & 25.16 & 41.48 & 45.01 \\
        UNet\cite{Unet} & 44.49 & 47.88 & 54.13 & 48.21 & 55.6 & 24.66 & 36.92 & 43.98 \\
        DeeplabV3+\cite{DeeplabV3++} & 46.28 & 50.68 & 49.07 & 51.53 & 60.79 & 26.27 & 40.23 & 45.39 \\
        Upernet\cite{UperNet} & 44.29 & 48.19 & 45.75 & 47.41 & 59.21 & 24.55 & 40.13 & 44.75 \\
        HRnet\cite{HRNet}  & 48.49 & 51.71 & 58.25 & 53.66 & 63 & 25.39 & 40.27 & 47.12 \\
        Swin-Upernet\cite{Swin} & 53.06 & 54.4 & 66.04 & 55.85 & 70.02 & 30 & \underline{45.56} & 49.59 \\
        SegFormer\cite{SegFormer}& 53.77 & 54.1 & 66.29 & 56.82 & 71.86 & 30.56 & 43.27 & 53.51 \\
        MANet\cite{MANet} & 50.72 & 53.68 & 63.37 & 53.86 & 66.45 & 29.14 & 40.79 & 47.73 \\
        DCSwin\cite{DCSwin} & 51.68 & 53.94 & 63.62 & \underline{57.83} & 68.36 & 24.81 & 44.54 & 48.66 \\
        Segnext\cite{Segnext}  & 52.56 & 54.78 & 65.46 & 57.9 & 66.9 & 28.54 & 39.22 & 55.15 \\
        FT-UNetFormer\cite{Unetformer} & 52.49 & 54.18 & 67.63 & 57.19 & 68.61 & 26.02 & 43.79 & 49.97 \\
        SFFNet\cite{SFFnet}  & 53.76 & 55.01 & \textbf{68.08} & 57.73 & \underline{72.73} & \underline{32.22} & 39.4 & 51.14 \\
        ConvLSR\cite{LSRFormer}  & \underline{54.72} & \underline{55.47} & \underline{67.65} & \textbf{58.24} & \textbf{72.94} & 31.34 & 40.95 & \underline{56.48} \\
		\hline
        FSDENet & \textbf{56.23} & \textbf{55.61} & 66.26 & 57.38 & 71.15 & \textbf{36.84} & \textbf{47.15} & \textbf{59.2} \\
    \bottomrule
	\end{tabular}
    \normalsize
    \label{table:loveda}
\end{table*}

MASF contributes the least to all metrics compared to the other modules, but MASF only contributes 0.9M parameters and 1.53G computations. The removal of MASF in \autoref{fig:FSDENet} leads to errors in the segmentation of smaller building classes, which may be because the edge texture information of smaller regions is more likely to be overwhelmed by the deeper semantic information during the process of feature fusion at multi-scale.

The HDEE module uses negligible parameters and 7.43G of computation but improves mF1 by 0.29 on the Vaihingen dataset and 0.33 on mIoU. Removing the HDEE module on FSDEnet also results in the most significant overall decrease in mF1 and mIoU scores. This is good evidence that our detail enhancement using the Haar wavelet transform significantly improves edge detail segmentation. It can also be seen from \autoref{fig:baseline} and \autoref{fig:FFDP} that HDEE plays a key role in the segmentation effect on the shadow region.

\subsection{Comparison With SOTA Models}
We compared the model's validity with recent SOTA methods on four widely used open-access datasets to verify the model's validity.

\begin{table*}[htbp]
	\centering
    \captionsetup{font={small}}
	\caption{COMPARISON WITH THE SOTA METHODS ON THE VAIHINGEN DATASET}
	\renewcommand{\arraystretch}{1.4} 
    \setlength{\tabcolsep}{11pt}
	\begin{tabular}{c|ccc|ccccc} 
        \toprule
    	\multirow{2}{*}{\textbf{Method}} & \multirow{2}{*}{\textbf{mF1}} & \multirow{2}{*}{\textbf{OA}} & \multirow{2}{*}{\textbf{mIoU}} & \multicolumn{4}{c}{\textbf{F1(\%)}}                               \\ 
    	\cline{5-9}&&&& \textbf{Imp. Surf}  & \textbf{Building} & \textbf{Low Veg.} & \textbf{Tree} & \textbf{Car}   \\ 	
		\hline
        UperNet\cite{UperNet} & 87.06 & 89.29 & 77.55 & 91.25 & 94.19 & 82.69 & 88.96 & 78.22 \\
        DeeplabV3+\cite{DeeplabV3++} & 87.02 & 88.95 & 77.42 & 91.07 & 93.79 & 82.52 & 88.44 & 79.29 \\
        HRnet\cite{HRNet} & 90.57 & 91.04 & 82.99 & 93.12 & 96.07 & 84.84 & 89.67 & 89.17 \\
        MANet\cite{MANet} & 89.36 & 90.02 & 80.95 & 91.51 & 94 & 83.82 & 90.15 & 87.34 \\
        SegFormer\cite{SegFormer}  & 90.38 & 91.01 & 82.7 & 93.43 & 96.13 & 84.45 & 89.38 & 88.5 \\
        FT-UNetFormer\cite{Unetformer} & 91.11 & 91.5 & 83.89 & 93.57 & 96.13 & 84.99 & 90.31 & 90.57 \\
        DCSwin\cite{DCSwin}  & 90.7 & 91.39 & 83.21 & 93.42 & 96.11 & 84.89 & 90.17 & 88.9 \\
        SegNext\cite{Segnext}  & 89.85 & 90.57 & 81.8 & 92.42 & 95.52 & 84.25 & 89.64 & 87.41 \\
        MPCNet\cite{mpcnet}  & 90.76 & 90.93 & 83.27 & 92.76 & 95.5 & 84.7 & 90.4 & 90.44 \\
        SFFNet\cite{SFFnet}  & 91.15 & 91.57 & 83.96 & 93.67 & 96.21 & 85.18 & 90.31 & 90.32 \\
        ConvLSR\cite{LSRFormer}  & \underline{91.35} & \underline{91.77} & \underline{84.29} & \underline{93.76} & \underline{96.31} & \underline{85.35} & \underline{90.56} & \underline{90.77} \\
        \hline
        FSDENet & \textbf{91.61} & \textbf{91.91} & \textbf{84.71} & \textbf{93.89} & \textbf{96.38} & \textbf{85.95} & \textbf{90.62} & \textbf{91.2} \\
    \bottomrule
	\end{tabular}
    \normalsize
    \label{table:vaihingen}
\end{table*}

\begin{table*}[htbp]
	\centering
    \captionsetup{font={small}}
	\caption{COMPARISON WITH THE SOTA METHODS ON THE POTSDAM DATASET}
	\renewcommand{\arraystretch}{1.4} 
    \setlength{\tabcolsep}{11pt}
	\begin{tabular}{c|ccc|ccccc} 
        \toprule
    	\multirow{2}{*}{\textbf{Method}} & \multirow{2}{*}{\textbf{mF1}} & \multirow{2}{*}{\textbf{OA}} & \multirow{2}{*}{\textbf{mIoU}} & \multicolumn{4}{c}{\textbf{F1(\%)}}                               \\ 
    	\cline{5-9}&&&& \textbf{Imp. Surf}  & \textbf{Building} & \textbf{Low Veg.} & \textbf{Tree} & \textbf{Car}   \\ 	
		\hline
        Fcn\cite{Fcn} & 91.44 & 89.82 & 84.47 & 92.43 & 95.77 & 85.93 & 88.04 & 95.04 \\
        UperNet\cite{UperNet} & 90.87 & 89.43 & 83.49 & 91.99 & 95.15 & 85.7 & 87.36 & 94.17 \\
        DeeplabV3+\cite{DeeplabV3++} & 88.848 & 89.18 & 83.44 & 91.82 & 94.69 & 85.27 & 87.99 & 84.47 \\
        HRnet\cite{HRNet} & 92.3 & 90.75 & 85.93 & 93.06 & 96.57 & 86.89 & 88.86 & 96.1 \\
        MANet\cite{MANet} & 89.29 & 87.56 & 80.93 & 89.68 & 92.61 & 81.62 & 88.11 & 94.41 \\
        SegFormer\cite{SegFormer} & 92.87 & 91.45 & 86.9 & 93.64 & 97.13 & 87.99 & 89.25 & 96.33 \\
        DCSwin\cite{DCSwin} & 93.01 & 91.68 & 87.13 & 93.68 & 97.13 & 88.51 & 89.49 & 96.24 \\
        Segnext\cite{Segnext} & 92.42 & 91.08 & 86.15 & 93.43 & 96.91 & 87.21 & 88.96 & 95.59 \\
        FT-UNetFormer\cite{Unetformer} & 92.97 & 91.42 & 87.07 & 93.2 & 96.9 & 88.56 & 89.57 & 96.64 \\
        MPCNet\cite{mpcnet} & 92.29 & 90.56 & 85.91 & 92.69 & 96.38 & 87.3 & 88.74 & 96.34 \\
        SFFNet\cite{SFFnet} & 93.11 & 91.63 & 87.3 & 93.58 & 97.11 & 88.57 & \textbf{89.74} & 96.55 \\
        ConvLSR\cite{LSRFormer} & \underline{93.23} & \underline{91.75} & \underline{87.54} & \underline{93.7} & \underline{97.34} & \underline{88.66} & 89.53 & \underline{96.93} \\
        \hline
        FSDENet & \textbf{93.35} & \textbf{91.87} & \textbf{87.73} & \textbf{93.78} & \textbf{97.37} & \textbf{88.81} & \underline{89.69} & \textbf{97.07} \\
    \bottomrule
	\end{tabular}
    \normalsize
    \label{table:potsdam}
\end{table*}
\begin{figure*}[htbp]
    \centering
    \includegraphics[width=\textwidth]{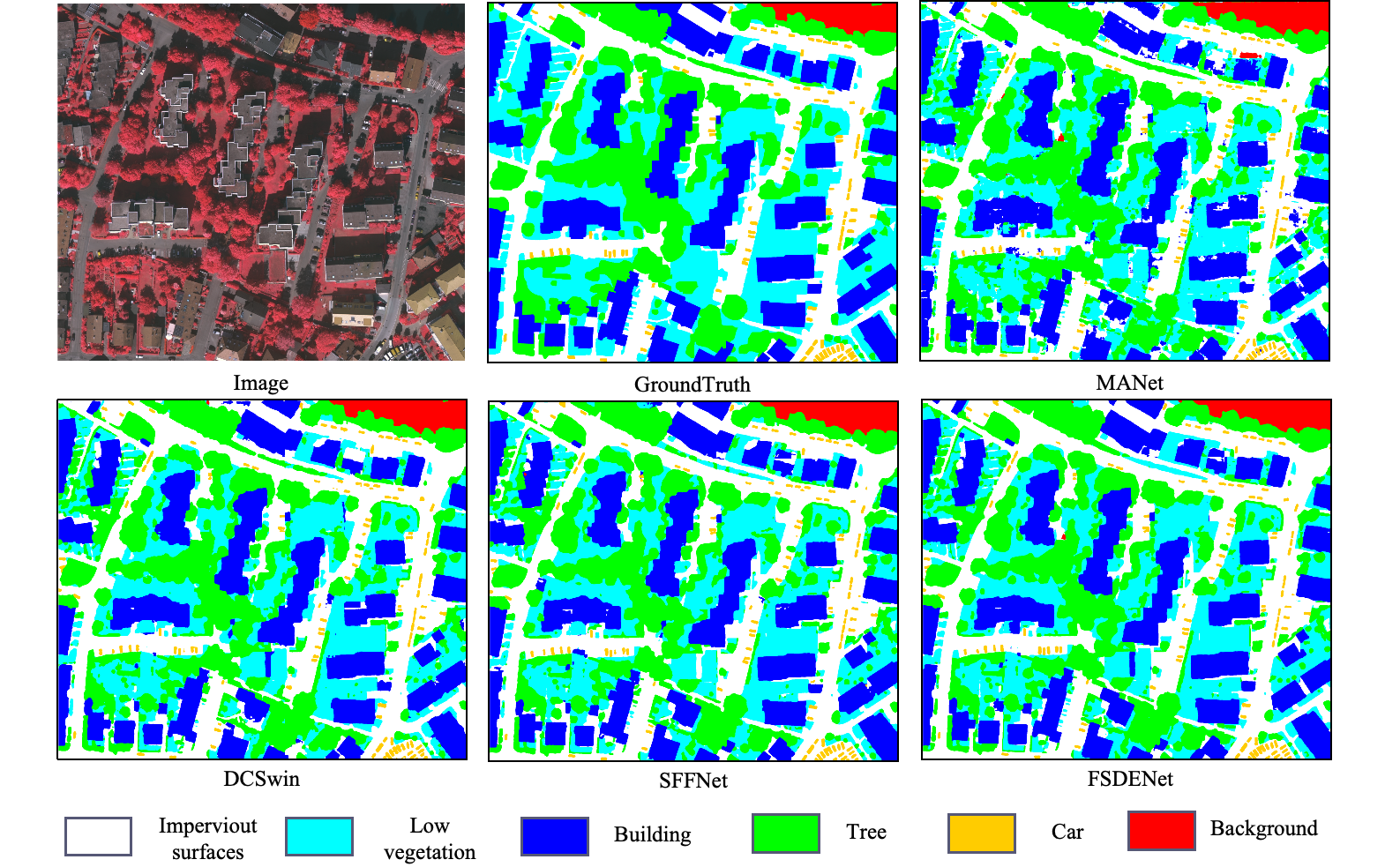}
    \captionsetup{font={small}}
    \caption{Qualitative comparisons between ours and other models on the Vaihingen dataset.}
    \label{fig:vaihingen}
\end{figure*}
\begin{figure*}[htbp]
    \centering
    \includegraphics[width=\textwidth]{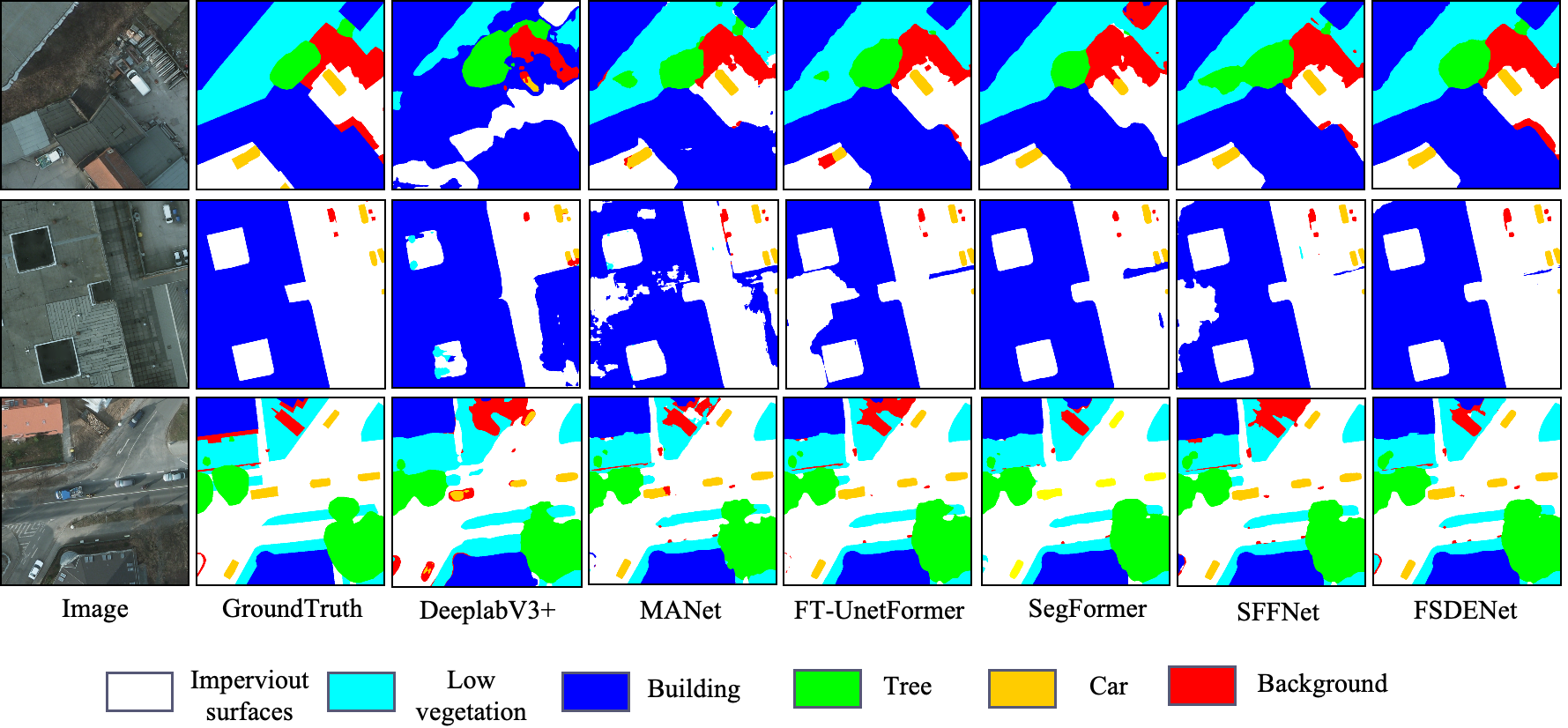}
    \captionsetup{font={small}}
    \caption{Qualitative comparisons between ours and other models on the Potsdam dataset.}
    \label{fig:potsdam}
\end{figure*}
1) Results on iSAID Dataset:
The main challenge of the iSAID dataset is the highly uneven distribution of foreground and background, as shown in \autoref{tab:isaid}. Our FSDENet achieves the same mIoU of 70.3 as the SegNeXt-L network with MSCAN-L as the backbone, but our model is less computationally intensive and complex. As shown in \autoref{fig:isaid}, in comparison with SegFormer-B4 using MiT-B4 as the backbone, our method segments the boundaries more thoroughly, and the first line shows that SegFormer-B4 fails to identify the swimming pool, while our method successfully does so, thanks to the excellent detail and boundary perception of our model.

2) Results on Vaihingen Dataset: The clutter/background category is included in our experiments but not reported. As seen from \autoref{table:vaihingen} and \autoref{table:parameters_flops}, our method achieves the highest mF1, mIoU, and OA. The FT-UNetFormer\cite{Unetformer} with a powerful Swin-Base\cite{Swin} backbone and the FSEENet improve the mF1 scores by 0.5\%, while our FLOPs and Params are only 60.94\% and 68.2\% of the FT-UNetFormer\cite{Unetformer}. In the previous SOTA model ConvLSR-Net\cite{LSRFormer}, which is also a ConvNeXt-small\cite{LSRFormer} backbone network, we are higher in all metrics, especially in the Car and Low Veg classes, where our method is higher by 0.57\% and 0.6\% respectively, due to the enhancement of the texture detail perception module and the frequency domain information from the well-designed texture detail perception module. Perception of the boundary and regions with small boundary changes.

3) Results on the Potsdam Dataset: The Potsdam dataset has the same categories as Vaihingen but with higher resolution, more texture detail, and more complex backgrounds and scenes, making it more challenging than Vaihingen. As shown in \autoref{table:potsdam}, our method achieves the best overall performance with scores of 93.35\% for mF1 and 91.87\% for OA. mIoU is 87.73. Compared to SFFNet, which also incorporates frequency domain information, our approach achieves higher results by 0.24\%, 0.25\%, and 0.53\% for mF1, OA, and mIoU, respectively, due to the targeted processing of different frequency domain information. 

\autoref{fig:vaihingen} shows the segmentation results of FSDENet on the image with ID 2 of the Vaihingen dataset. Meanwhile, as shown in \autoref{fig:potsdam}. In the comparison graph of locally zoomed images on the Potsdam dataset, FSDENet exhibits better segmentation ability when dealing with complex backgrounds, particularly for more challenging clutter and backgrounds. As in the first row, both our method and SFFNet, which also incorporates frequency domain information, segment the clutter/background next to the house, whereas our method segments it completely. In contrast, the other methods do not segment it. FT-UNetFormer and SegFormer both misclassify the car part as clutter/background. The second and third rows demonstrate that our method can better handle spatial correlation and boundaries.
\begin{figure*}[htbp]
    \centering
    \includegraphics[width=\textwidth]{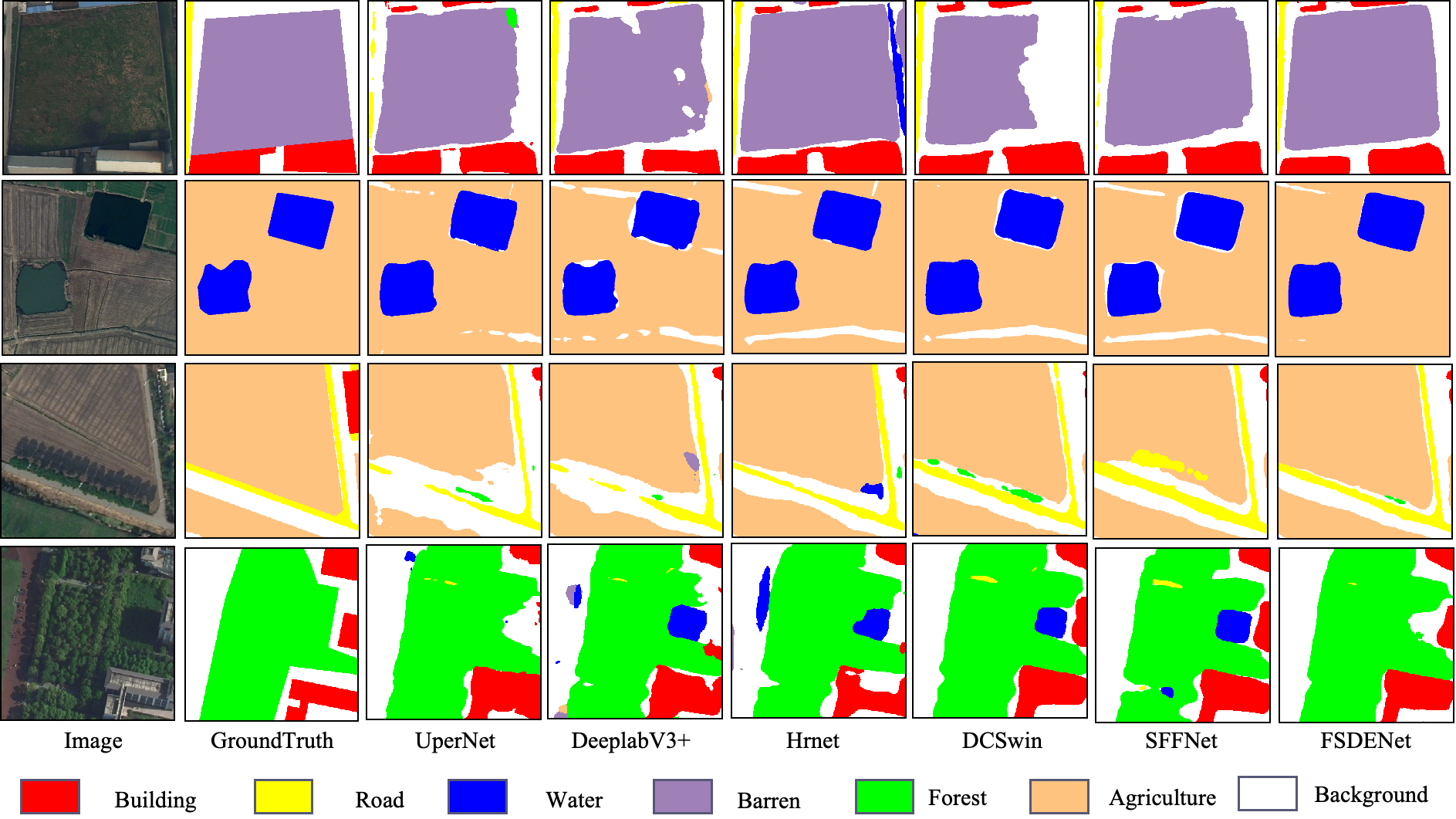}
    \caption{Qualitative comparisons between ours and other models on the LoveDA dataset}
    \label{fig:loveda}
\end{figure*}

4) Results on LoveDA Dataset:
LoveDA is a large-scale land cover segmentation dataset. As shown in \autoref{table:loveda}, FSEENet significantly outperforms the existing models in the main metric mIoU. Moreover, the mIoU is 2.47\% and 1.51\% higher than the recently proposed SOTA models SFFNET and ConvLSR-Net, respectively. Notably, FSEENet is 5.5\% higher than the existing methods in the Barren category, 6.2\% higher in the Forest category, and 2.72\% higher in the agricultural category because we effectively introduce the frequency domain information. Our method has better accuracy in dealing with the regions where the changes in the edge contour are not noticeable, such as trees, cultivated land, etc.

Objects in the same category may appear in different shapes, textures, and colors. As shown in \autoref{fig:loveda}, for example, the agricultural category in the third row is covered by shadows, resulting in internal texture changes, leading to segmentation errors in DeeplabV3+\cite{DeeplabV3++}, SFFNet\cite{SFFnet}, etc., whereas our method has better segmentation results due to its stronger perception of grey-scale changing regions. The same texture features can also appear in different classes; for example, other methods segment the foster category as building in the fourth row, while our method does not miss segmenting it. The segmentation results in the first and second rows also demonstrate that our method is highly accurate in handling boundary information.

\section{DISSCUSION}
In this study, we propose the FSDENet network, which demonstrates outstanding performance across multiple publicly available remote sensing semantic segmentation datasets. In particular, the model excels in challenging scenarios such as shadow occlusion, low-contrast regions, and blurred object boundaries. By effectively integrating spatial and frequency domain information, FSDENet enhances edge perception and detail reconstruction from multiple perspectives, significantly improving segmentation accuracy in semantic boundary and fine-detail regions.

Although each individual module contributes differently to the final performance, the collaborative effect of the four modules leads to the most substantial overall improvement. Specifically, as shown in \autoref{fig:gam}, the HWDE module exhibits the most prominent effect in restoring boundary details, while the FFDP module excels in enhancing the model’s sensitivity to grayscale transitions. In contrast, the MASF module contributes a relatively smaller improvement to the overall mIoU; however, it plays an indispensable role in the fine-grained segmentation of detailed regions. On the LoveDA dataset, the spectral features of the agriculture class are influenced by crop type and growth stage, often causing confusion with the barren and forest categories. Experimental results show that FSDENet significantly outperforms existing methods on these three classes, which we attribute primarily to the incorporation of frequency-domain features that enhance the model’s ability to perceive grayscale variations and subtle boundary cues.

Despite the strong performance of FSDENet across various datasets, several limitations remain. For example, on the LoveDA dataset, our model shows relatively lower performance in identifying road and water classes compared to some existing approaches. This may be due to a bottleneck in the frequency-domain components when handling elongated structures with extensive spatial continuity, presenting a new challenge for future improvements in frequency-domain modeling. Additionally, the stability of frequency-domain feature extraction may be affected under extremely complex backgrounds or high-noise conditions. Furthermore, although each module is designed with a clear functional division, the overall network architecture is more complex than traditional methods, and its applicability in edge deployment or lightweight scenarios requires further optimization.

\section{CONCLUSION}

Since local and global context information is crucial for the semantic segmentation of aerial images, this paper proposes a method that leverages CNN to extract multi-scale local features. To preserve high-resolution, detailed texture information, we unify the scale size of these extracted features. The Multi-Attention Select Fusion (MASF) block is employed to align the receptive fields of features across different scales, ensuring that shallow detailed texture information is not overwhelmed by deep global semantic information. The Cross-Agent Global Fusion (CAGF) block utilizes cross-agent attention to complement global details. At the same time, agent tokens reduce computational complexity during information interaction between features, further refining receptive field alignment. To effectively incorporate frequency domain information, the Fast Fourier Detail Perception (FFDP) block employs extensive kernel decomposition to complement global information and enhance feature diversity through multiple large kernel convolutions. The fast Fourier transform is also utilized to introduce frequency domain information into the international context, improving the model's perception of grayscale variations. The Haar Wavelet Detail Enhancement (HWDE) module decomposes the original image into high and low-frequency signals using Haar wavelet downsampling to refine segmentation accuracy for detailed textures further. It exploits these properties to enhance the model's detail perception and edge segmentation capabilities.

Extensive experimental results demonstrate that our method, which effectively fuses spatial local information, spatial global information, and frequency domain information, enhances the model's ability to address issues of low contrast and edge semantic ambiguity caused by grayscale changes due to occlusion and shading. Our model achieves state-of-the-art (SOTA) results on four open aerial image segmentation datasets while maintaining excellent model complexity.

\bibliographystyle{IEEEtran}
\bibliography{reference}

% 作者1

\begin{IEEEbiography}[{\includegraphics[width=1in,height=1.25in,clip,keepaspectratio]{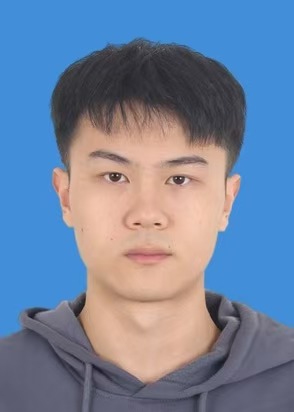}}]{Jiahao Fu}
received his B.S. in Software Engineering from Xinjiang University in Urumqi, China, in 2022. He is pursuing an M.S. degree at Xinjiang University, Urumqi, China.
His research interests include computer vision and remote sensing semantic segmentation.

\end{IEEEbiography}

\vspace{5pt}
% 作者2
\begin{IEEEbiography}[{\includegraphics[width=1in,height=1.25in,clip,keepaspectratio]{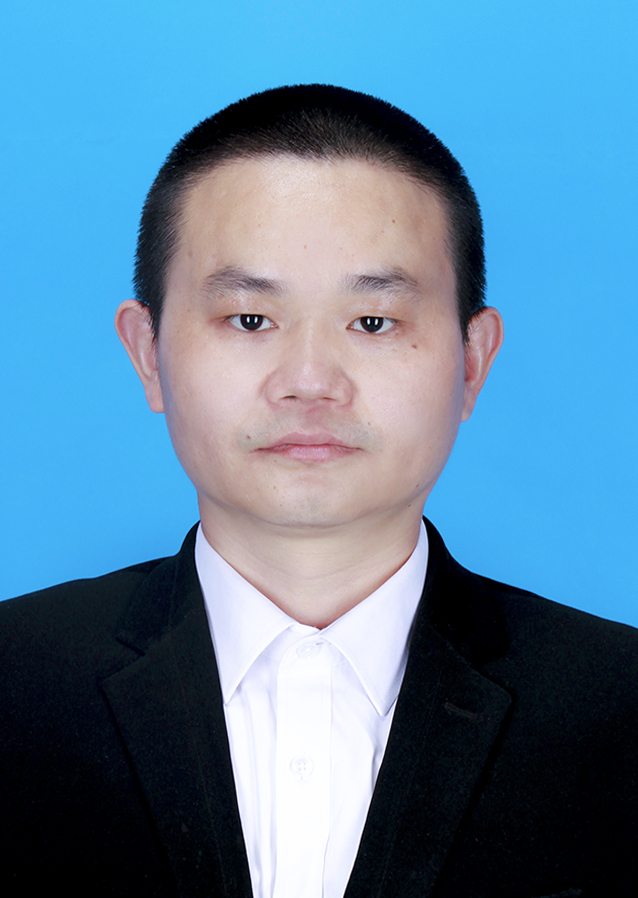}}]{Yinfeng Yu}
(Member, IEEE) received the Ph.D degree from the School of Computer Science and Technology, Tsinghua University, Beijing, China, in 2023. He is an Associate Professor at the School of Computer Science and Technology, Xinjiang University.
His research interests include embodied AI, computer vision, multimodal processing, and remote sensing information processing.
\end{IEEEbiography}

\vspace{5pt}

% 作者3
\begin{IEEEbiography}[{\includegraphics[width=1in,height=1.25in,clip,keepaspectratio]{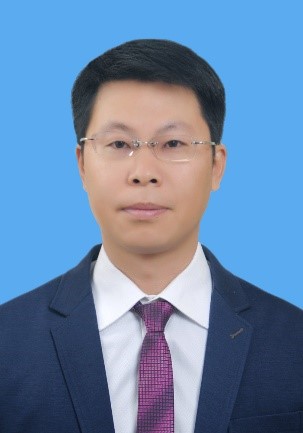}}]{Liejun Wang}
received the Ph.D. from the School of Information and Communication Engineering, Xi'an Jiaotong University, Xi'an, China, 2012.
He is now a Professor at the School of Computer Science and Technology, Xinjiang University, Urumqi, China. His research interests include wireless sensor networks, computer vision, and natural language processing. % 作者介绍内容
 
\end{IEEEbiography}

\vfill
 
\end{document}